\documentclass[journal]{IEEEtran}
\usepackage{algpseudocode}
\usepackage{amsmath,amssymb,amsfonts}
\usepackage{amsthm}
\usepackage{algpseudocode}
\usepackage{algorithm}
\usepackage{mathrsfs}
\usepackage{xcolor}

\def\BibTeX{{\rm B\kern-.05em{\sc i\kern-.025em b}\kern-.08em
    T\kern-.1667em\lower.7ex\hbox{E}\kern-.125emX}}
\usepackage{array}
\usepackage[caption=false,font=normalsize,labelfont=sf,textfont=sf]{subfig}
\usepackage{textcomp}
\usepackage{stfloats}
\usepackage{url}
\usepackage{verbatim}
\usepackage{graphicx}
\usepackage{cite}
\usepackage{multirow}

\begin{document}
\title{Toward Semantic Communication for Real-time Mobile 3D Reconstruction}

\author{Fangzhou Zhao, Yao Sun$^{\star}$,~\IEEEmembership{Senior Member,~IEEE}, Xuesong Liu,~\IEEEmembership{Graduate Student Member,~IEEE}, Runze Cheng~\IEEEmembership{Member,~IEEE}, Shang Kai and Yi Sun
\thanks{Fangzhou Zhao and Yi Sun are with the School of Electrical and Electronic Engineering, North China Electric Power University, Beijing 102206, China.}
\thanks{Yao Sun, Xuesong Liu and Runze Cheng are with the James Watt School of Engineering, University of Glasgow, Glasgow G12 8QQ, U.K.}
\thanks{Shang Kai is with the School of Computer Science and Technology, China University of Petroleum (East China), Qingdao 257061, China.}
\thanks{$^\star$Corresponding author. (Email: Yao.Sun@glasgow.ac.uk)
}}

\maketitle

\begin{abstract}
Real-time mobile 3D reconstruction is fundamental to many emerging applications such as autonomous navigation and digital twin construction, where a moving platform continuously captures an image stream and transmit to a computing server for scene understanding.
Unlike offline reconstruction, camera poses and scene geometry are estimated on-the-fly during acquisition, making multi-view consistency a real-time requirement and rendering geometric estimation highly sensitive to communication-induced distortions.
Semantic communication (SemCom) transmits compact semantic information, offering a promising way to preserve task-critical data over unreliable links. However, existing designs are optimized at the image or single-view level and without providing explicit reliability information for geometric estimation, limiting their applicability to real-time mobile 3D reconstruction.
In this context, we propose a SemCom framework for real-time mobile 3D reconstruction. The framework includes a semantic transceiver that outputs a reconstructed image alongside a pixel-wise confidence map, quantifying the reliability of each region. We further introduce a confidence-guided geometric estimation method, incorporating confidence into RANSAC-based pose initialization and bundle adjustment to reduce the influence of unreliable regions and enhance robustness under noisy channels. Simulations show that, compared to existing SemCom and traditional seperate source and channel coding, our framework maintains high image quality while significantly improving pose estimation accuracy and 3D structural consistency.
\end{abstract}


%
\IEEEpeerreviewmaketitle

\section{Introduction}
Real-time mobile 3D reconstruction plays a central role in many vision-based applications, such as autonomous navigation, digital twin construction, immersive augmented and virtual reality, and remote inspection \cite{realtime1,realtime4}. In typical deployment scenarios, a mobile terminal agent (such as a robot, an unmanned platform, or a handheld device) continuously captures image sequences of the environment, which are transmitted to a computing server for 3D reconstruction \cite{realtime2,realtime3}.
In such mobile reconstruction tasks, camera motion and scene geometry should be incrementally inferred during the acquisition process, rather than being analyzed after all images have been collected, which fundamentally distinguishes real-time mobile 3D reconstruction from offline 3D reconstruction \cite{realtime1}.
By leveraging multi-view geometric relationships across consecutive frames, mobile 3D reconstruction applications estimate camera motion and recover sparse or dense scene structure in real-time \cite{3DGS,colmap}. The accuracy of this process critically depends on the consistency of visual information across views, since errors in feature correspondences or camera pose estimation can propagate through the reconstruction pipeline and degrade the recovered geometry. As a result, real-time 3D reconstruction imposes strict requirements on the quality and geometric reliability of the visual data used for multi-view estimation.

In practical mobile settings, the visual data captured by the mobile terminal agents is typically transmitted to a remote or edge computing server over wireless links. Such transmission is inherently subject to bandwidth limitations and channel conditions, which introduce distortions and uncertainty into the received image sequence \cite{mobile1,mobile2}. For real-time 3D reconstruction, these effects are particularly harmful because the reconstruction process depends on maintaining consistent visual information across multiple views. Even small local distortions can make visual information less consistent across frames, which in turn leads to inaccurate motion estimation and unstable reconstruction \cite{realtime1}. If these inconsistencies propagate through the multi-view geometric estimation process, their accumulation across views results in noticeable geometric drift and degraded reconstruction quality. Consequently, uncertainty introduced during image transmission directly degrades the reliability of multi-view estimation and poses a fundamental challenge to robust and accurate real-time mobile 3D reconstruction.

In this context, semantic communication (SemCom) offers a promising solution to the aforementioned communication challenges. Unlike conventional communication schemes that aim to preserve bit-level or pixel-level fidelity, SemCom focuses on transmitting compact semantic representations that capture information most relevant for downstream processing. \cite{sc2,sc3,xueosng1}. This task-oriented design aligns well with the requirements of 3D reconstruction, where maintaining geometric consistency and stable structural information across multiple views is typically more important than producing visually plausible individual recovered images. By prioritizing semantically meaningful visual information over pixel-level accuracy, SemCom provides a potential pathway toward improving the robustness of 3D reconstruction under unreliable communication conditions.

Despite this potential, existing SemCom approaches are not directly suited for real-time 3D reconstruction.
Most current designs treat SemCom as preserving the semantic content of individual images and therefore optimize single-view reconstruction quality or perception performance \cite{sc1,scc,runze}.
However, in 3D reconstruction, the value of semantic information is not determined by the visual plausibility but by how reliably it supports the incremental construction of a consistent 3D structure across multiple views. 
Visually plausible yet geometrically inconsistent regions can be as detrimental as severely distorted ones, whereas visually imperfect regions that preserve stable structural cues may be more valuable for accurate 3D estimation.
Treating all reconstructed regions equally prevents geometric estimation from properly accounting for observation reliability.
This issue is further amplified in real-time mobile reconstruction.
Unlike offline scenarios, where unreliable observations can be revisited or corrected through global optimization, real-time reconstruction proceeds incrementally as the platform moves, and each frame is immediately integrated into pose estimation and structure recovery.
As a result, unreliable semantic content may be irreversibly embedded into the evolving 3D model and propagate across subsequent views, leading to accumulated geometric errors.
Therefore, SemCom for real-time 3D reconstruction should take 3D geometry usability as the core objective instead of individual image semantic quality, ensuring structural consistency under multi-view conditions and directly serving feature matching, pose estimation, and multi-view optimization.

To address the above challenges, this paper proposes a novel SemCom framework for real-time mobile 3D reconstruction. Specifically, we develop a SemCom transceiver that delivers visual information tailored for 3D reconstruction. In addition to reconstructed images, the transceiver provides explicit reliability information about the reconstructed visual information, which is subsequently incorporated into camera pose estimation and multi-view optimization. By allowing communication-aware reliability information to directly influence geometric estimation, the proposed framework improves the robustness and consistency of 3D reconstruction under imperfect communication conditions.

The main contributions of this work can be summarized as follows:
\begin{itemize}
    \item 
    \textit{SemCom framework for real-time mobile 3D reconstruction:}  
    In the proposed framework, a mobile terminal captures image sequences of the environment and transmits them to a receiver via SemCom, where visual information relevant to multi-view geometry is prioritized to support camera pose estimation and 3D reconstruction under imperfect communication conditions. In 3D reconstruction process, the pose estimation component is further designed to explicitly incorporate information produced by semantic transmission, enabling the geometric reconstruction process to account for communication-induced uncertainty.

    \item \textit{SemCom transceiver design for 3D reconstruction:}  
    In this transceiver, the decoder is explicitly designed to jointly output reconstructed images and a pixel-wise confidence map, where the confidence map characterizes the reliability of image reconstruction under imperfect communication conditions. This transceiver provide both visual information and reconstruction confidence to the 3D reconstruction process, providing essential inputs for confidence-guided geometric estimation.

    \item \textit{Confidence-guided geometric estimation scheme:}
    Building on the pixel-wise confidence maps provided by the SemCom transceiver, the proposed geometric estimation scheme incorporates reconstruction confidence into pose initialization and multi-view optimization, allowing reliable observations to be emphasized while reducing the influence of unreliable regions. This design enables geometric estimation to explicitly account for communication and image reconstruction induced uncertainty, improving the robustness and consistency of 3D reconstruction under imperfect communication conditions.

    \item \textit{Numerical simulation validation:}
    We validate the proposed framework through simulations on 3D reconstruction and image transmission.
    The proposed SemCom-enabled 3D reconstruction method is integrated into a standard multi-view geometry pipeline and evaluated under imperfect communication conditions. Simulation results demonstrate that incorporating SemCom and confidence-guided geometric estimation improves the robustness and consistency of camera pose estimation and 3D reconstruction compared with conventional transmission-based approaches.

\end{itemize}

The rest of this paper is organized as follows. 
Section II reviews the related work. 
Section III introduces the proposed SemCom framework for real-time mobile 3D reconstruction. 
Section IV presents the design of the SemCom transceiver design for 3D reconstruction. 
Section V describes the proposed confidence-guided camera pose estimation and multi-view optimization methods. 
Section VI reports the numerical simulation results.
Finally, Section VII concludes the paper.

\section{Related Work}
With the growing deployment of mobile 3D reconstruction systems, reliable visual data transmission has become an integral part of practical reconstruction pipelines. In wireless and mobile settings, bandwidth limitations and dynamic channel conditions introduce additional challenges to maintaining reconstruction accuracy and stability. In parallel, recent advances in SemCom have provided new perspectives on information transmission. In recent years, researchers have made significant progress in both mobile 3D reconstruction and SemCom. This section reviews related research on these two aspects separately.

\subsection{Real-time Mobile 3D Reconstruction}
Recent advances in real-time mobile 3D reconstruction have led to significant progress in camera pose estimation, scene representation, and large-scale environment modeling. A wide range of studies have investigated efficient and robust reconstruction pipelines that operate on image sequences captured by mobile platforms, demonstrating improved accuracy and scalability in dynamic and complex environments.

For instance, the study of \cite{realtime1} presents a neural network-based framework that reconstructs 3D scenes in real-time from monocular video by using a network to sequentially build local surface patches, integrating them with a learned fusion module. The authors in \cite{moblie3D1} describes a real-time monocular dense SLAM system, built upon the MASt3R prior, that operates without a fixed camera model to produce globally consistent poses and dense geometry from in-the-wild videos at a frame rate of 15. The study of \cite{moblie3D2} introduces edge SLAM, an edge-assisted semantic visual SLAM service for mobile devices that improves localization accuracy and achieves real-time performance through efficient computation offloading. The work of \cite{moblie3D3} proposes LAMV-SLAM, a LiDAR-assisted monocular visual SLAM framework for outdoor mobile robots that achieves real-time, real-scale dense mapping without GPU by fusing LiDAR and visual data.

While these studies have achieved impressive performance in real-time mobile 3D reconstruction, they typically assume reliable and high-quality visual inputs at the reconstruction backend. 
In practical mobile deployments, however, visual data often need to be transmitted over wireless links, where communication imperfections can introduce distortions and uncertainty that directly affect reconstruction accuracy. 
As a result, existing real-time 3D reconstruction methods are not explicitly designed to account for communication-induced uncertainty, limiting their robustness in communication-constrained mobile scenarios.

\subsection{SemCom for Wireless Visual Transmission}
Motivated by the challenges of communication-limited perception systems, recent studies in SemCom have explored transmission strategies for wireless visual data.

For example, the study of \cite{scsc1} proposes an integrated sensing, computing, and SemCom framework for vehicular networks, which jointly optimizes semantic secrecy rate and sensing accuracy under computing constraints via Bernstein-type inequality and alternating optimization techniques.
The research in \cite{scsc2} proposes a SemCom framework to enhance UAV-UGV cooperative path planning in unreliable wireless conditions by transmitting only key semantic information, which reduces data volume while maintaining accuracy.
The work in \cite{scsc3} describes a SemCom framework for lunar lander-to-satellite image transmission that dynamically adjusts the transmission strategy based on real-time control feedback to ensure reliable delivery of critical features for autonomous landing.
The study of \cite{scsc4} proposes U-DeepSC, a unified SemCom system that serves multiple tasks using a dynamic feature transmission scheme and a shared codebook to reduce overhead and model size while maintaining performance comparable to task-specific systems.
The study of \cite{chengsi} proposes KG-SemCom, a SemCom framework that integrates knowledge graphs to enhance image encoding and decoding by fusing contextual and structured knowledge for improved accuracy and robustness.

Despite these advances, existing SemCom methods are not directly applicable to real-time mobile 3D reconstruction. Most current approaches are designed for specific perception or control tasks and focus on improving transmission efficiency or task performance, without explicitly considering the tight coupling between communication uncertainty and geometric reconstruction processes. As a result, these methods do not fully address the requirements of mobile 3D reconstruction, where communication-induced distortions can directly affect camera pose estimation and multi-view geometric consistency.

\section{System Model}
\begin{figure*}[htp]
	\centering
        \setlength{\abovecaptionskip}{-0pt}
        \setlength{\belowcaptionskip}{-0pt}
	\includegraphics[width=0.95\textwidth]{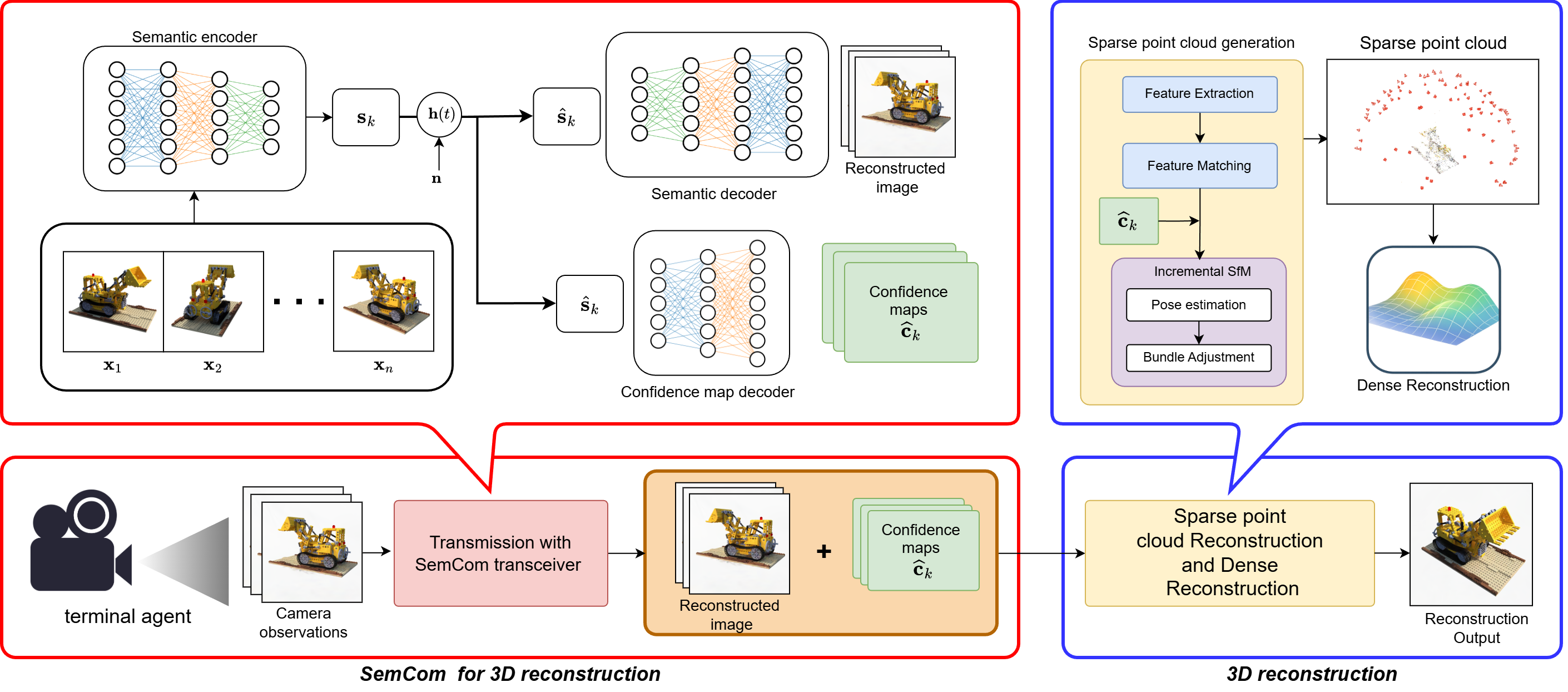}
  	\caption{Overview of the proposed SemCom framework for real-time mobile 3D reconstruction, where the red module illustrates the SemCom process and the blue module depicts the subsequent 3D reconstruction pipeline.}
	\label{img2}
\end{figure*}

Consider a real-time mobile 3D reconstruction system comprising a mobile terminal agent with a monocular camera and a computing server for 3D reconstruction. 
As shown in Fig. 1, the mobile terminal agents first take a sequence of images through the monocular camera, and at the same time transmits the real-time images to the computing server through a noisy channel by the semantic transmitter. Then, the computing server performs 3D reconstruction based on the received images.
In the process of 3D reconstruction, the pose estimation is first performed on the real-time received image, and then the sparse point cloud is calculated according to the obtained image and pose.

\subsection{SemCom Model}
We exploit a pair of semantic transceivers to accurately transmit the image through the noisy channel from the mobile terminal agents to the computing server. 
For a 3D reconstruction task, the ordered set of all images taken by its mobile terminal agents in sequence is denoted as $\mathcal X=\{\mathbf x_1, \mathbf x_2, \cdots, \mathbf x_n\}$, where $\mathbf x_k$ denotes the $k$th image.
In the transmission of $\mathbf x_{k}$ based on SemCom, $\mathbf x_{k}$ is first encoded into bits representing semantic $\mathbf s_{k}$ by a semantic encoder, denoted as:
\begin{equation} 
\mathbf s_{k} = {\mathcal{ E}_e}\left ({{\mathbf {x}_{k}}; { \boldsymbol \alpha}} { }\right),
\end{equation}
where ${\mathcal{ E}_e}$ is the neural networks of the encoder with the trainable parameters $\boldsymbol \alpha$.
After $\mathbf s_{k}$ are transmitted through the noisy physical channel, the received bits $\mathbf {\hat s}_{k}$ are
\begin{equation} 
\mathbf {\hat s}_{k} = {\mathbf h(t) *\mathbf s_{k} + {\mathbf {n}}},
\end{equation}
where $\mathbf h(t)$ denotes the channel coefficient, and $\mathbf {n}$ is the additive white Gaussian noise (AWGN) of the channel.
At the receiver, $\mathbf {\hat s}_{k}$ is reconstructed into an image $\mathbf {\hat x}_{k}$ by the corresponding semantic decoder, represented as
\begin{equation} 
{\mathbf {\hat x}}_{k} = {\mathcal{ E}_d}\left ({\mathbf {\hat s}_{k}; { \boldsymbol \alpha'}} {}\right),
\end{equation}
where ${\mathcal{ E}_d}$ is the neural networks of the semantic decoder with the trainable parameters $\boldsymbol \alpha'$.

Since SemCom relies on learned reconstruction at the receiver, distortions introduced by channel noise may lead to locally unreliable or inaccurately reconstructed image regions. Such reconstruction uncertainty can adversely affect downstream geometric estimation if all visual information are treated equally. This motivates the need to analyze and quantify the confidence of reconstructed images, so as to mitigate the impact of unreliable regions on subsequent 3D reconstruction.
In particular, while decoding, a confidence estimation module trained simultaneously with the semantic transceiver gives a confidence map $\mathbf {\hat c}_{k}$ according to $\mathbf {\hat s}_{k}$, denoted as:
\begin{equation} 
{\mathbf {\hat c}}_{k} = {\mathcal{ E}_c}\left ({\mathbf {\hat x}_{k}; { \boldsymbol \beta}} {}\right),
\end{equation}
where ${\mathcal{ E}_c}$ is the confidence estimation networks with the trainable parameters $\boldsymbol \beta$ and $\mathbf {\hat c}_{k}$ is a matrix with the same length and width as $\mathbf {\hat x}_{k}$. Each element of the matrix corresponds to the confidence of the corresponding pixel in $\mathbf {\hat x}_{k}$.

\subsection{3D Reconstruction Model}
In the computing server, we first exploit pose estimation according to the reconstructed image, its presequence images and the corresponding confidence maps. Then, the computing server performs pose estimation based on the reconstructed images and their corresponding confidence maps. 

Since the $k$th and $(k+1)$th reconstructed images are captured from two adjacent viewpoints of the same scene, their geometric relationship can be represented by a rigid-body transformation defined by a rotation matrix $\mathbf{R}_{k,k+1}$ and a translation vector $\mathbf{T}_{k,k+1}$. 
Accordingly, the pose estimation module can be generally described as a mapping function
\begin{equation}
(\mathbf{R}_{k,k+1}, \mathbf{T}_{k,k+1}) 
= \mathcal{F}_{\text{pose}}(\mathbf{\hat{x}}_{k}, \mathbf{\hat{x}}_{k+1}, \mathbf{\hat{c}}_{k}, \mathbf{\hat{c}}_{k+1}),
\end{equation}
where $\mathcal{F}_{\text{pose}}(\cdot)$ denotes the pose estimation algorithm that infers the relative camera transformation between two reconstructed images according to their visual and confidence information. 
This module provides the relative pose set $\{\mathbf{R}_{k,k+1}, \mathbf{T}_{k,k+1}\}$ for all adjacent image pairs in the sequence, serving as the geometric foundation for subsequent 3D reconstruction.

After obtaining the relative poses, the global camera poses $\{\mathbf{R}_{k}, \mathbf{T}_{k}\}_{k=1}^{n}$ are derived through composition across the sequence. 
Based on the reconstructed images and their estimated poses, the system generates a sparse 3D point cloud through triangulation. 
For a calibrated camera with intrinsic matrix $\mathbf K$, the projection relationship between a 3D point $\mathbf{P}_i=[X_i,Y_i,Z_i,1]^{\top}$ 
and its projection $\mathbf{p}_i^k=[u_i^k,v_i^k,1]^{\top}$ in the $k$-th image can be expressed as
\begin{equation}
\lambda_i^k \mathbf{p}_i^k = \mathbf K [\mathbf{R}_k|\mathbf{T}_k]\mathbf{P}_i,
\end{equation}
where $\lambda_i^k$ is the depth-related scale factor. 
By solving multiple projection constraints across the image sequence, a sparse 3D point set can be obtained as
\begin{equation}
\mathcal{P} = \mathcal{G}(\mathbf{\hat{x}}_{1:n}, \mathbf{R}_{1:n}, \mathbf{T}_{1:n}, \mathbf{\hat{c}}_{1:n})
= \{\mathbf{P}_1,\mathbf{P}_2,\dots,\mathbf{P}_N\},
\label{eq:structure}
\end{equation}
where $\mathcal{G}(\cdot)$ denotes the general triangulation or structure-from-motion mapping function for reconstructing the sparse 3D structure from the image sequence, estimated poses, and confidence maps.

The mathematical formulation presented here does not correspond to a specific reconstruction algorithm but instead provides a general form applicable to a class of methods that estimate camera poses and reconstruct sparse 3D structures based on multi-view geometric constraints.

\section{SemCom Transceiver for 3D Reconstruction}
Based on the proposed semantic transfer framework, we design a SemCom transceiver, named 3DC-SC, to transmit images and simultaneously generate pixel-wise confidence maps of the images. 
The 3DC-SC transceiver employs a convolutional encoder–decoder structure, containing an encoder and two decoders, which output a restored image and a confidence map, respectively. The confidence prediction network that estimates the reliability of each reconstructed pixel to guide the pose estimation for the 3D reconstruction.
\begin{figure*}[htp]
	\centering
        \setlength{\abovecaptionskip}{-0pt}
        \setlength{\belowcaptionskip}{-0pt}
	\includegraphics[width=0.95\textwidth]{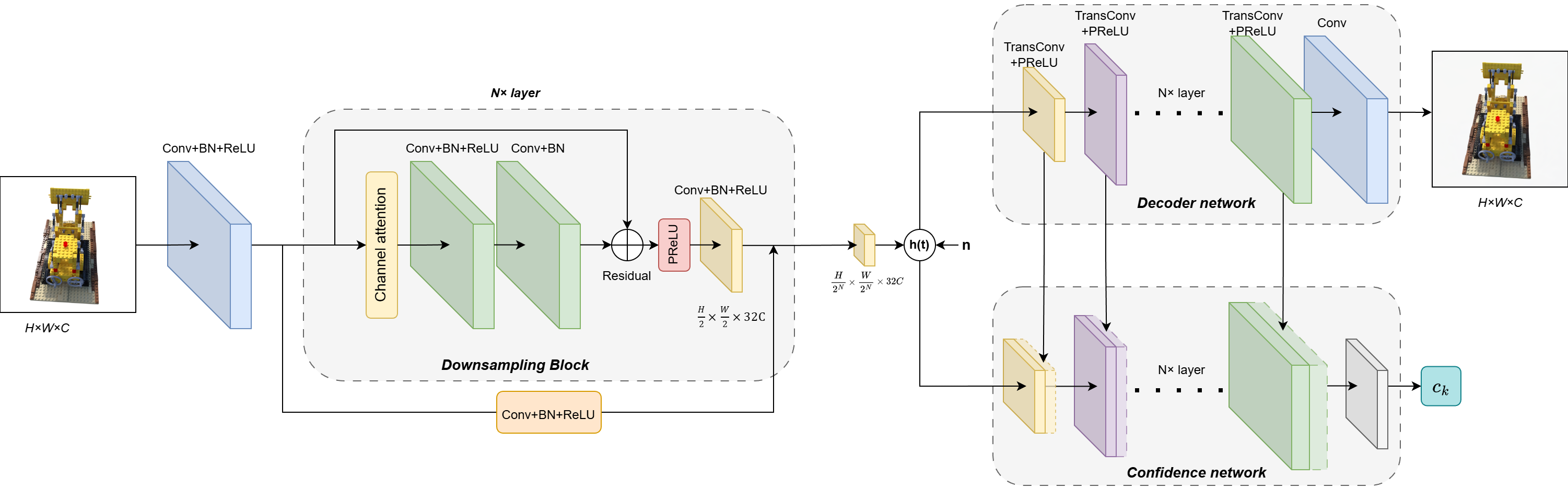}
  	\caption{The proposed 3DC-SC transceiver with a convolutional encoder and dual decoders, where visual features are encoded into semantic representations, transmitted over a noisy channel, and decoded into a reconstructed image and a pixel-wise confidence map.}
	\label{img2}
\end{figure*}

\subsection{Structure of the Semantic Encoder and Decoder}
As shown in Fig.~2, we design a SemCom transceiver for image transmission in 3D reconstruction tasks.
The proposed transceiver adopts a convolutional encoder--decoder architecture.
Given an input image $\mathbf x_k \in \mathbb{R}^{H \times W \times C}$, the encoder first applies a convolutional feature extraction layer combined with batch normalization and nonlinear activation to obtain low-level visual features capturing local textures and edges.
Subsequently, the encoder employs a set of convolutional encoding modules to further model the extracted features.
Each encoding module consists of a channel attention mechanism \cite{se} and a convolutional residual unit.
Specifically, the channel attention module exploits global contextual statistics to adaptively reweight feature channels, thereby emphasizing semantic features that are more critical for geometric structure preservation and local consistency.
At the end of each encoding module, a downsampling operation is introduced to progressively reduce the spatial resolution and form a compact feature representation.
After the encoding process, the input image is transformed into a latent feature tensor $z_k \in \mathbb{R}^{\frac{H}{2^L} \times \frac{W}{2^L} \times C_L}$,
where \(L\) denotes the number of downsampling stages and \(C_L\) represents the number of channels at the bottleneck layer.

At the receiver, the decoder mirrors the encoder with a symmetric convolutional architecture.
A sequence of upsampling operations implemented via transposed convolutions is applied to gradually restore the spatial resolution of the latent features.
At each decoding stage, convolutional layers and nonlinear activations are employed to refine the recovered features and alleviate distortions caused by imperfect data transmission.
Finally, a convolutional mapping layer reconstructs the output image $\hat{x}_k \in \mathbb{R}^{H \times W \times C}$.
The reconstructed image \(\hat{x}_k\) is subsequently fed into the 3D reconstruction pipeline for feature extraction, correspondence establishment, and camera pose estimation.

To jointly optimize pixel-wise fidelity and geometric recoverability, we adopt a combined loss function that includes a reconstruction term and a descriptor consistency term.

The reconstruction loss is defined as the mean squared error (MSE) between the reconstructed image and the original input:
\begin{equation}
\mathcal{L}_{\text{MSE}} = \frac{1}{N} \sum_{k=1}^{N} \| x_k - \hat{x}_k \|_2^2.
\label{eq:mse}
\end{equation}

To enhance the structural consistency of keypoint-level features, we introduce a descriptor consistency loss. For each pair of original and reconstructed images, local descriptors are extracted and matched using nearest-neighbor correspondence. Let \( \{(f_i^k, \hat{g}_i^k)\}_{i=1}^{K} \) denote the matched descriptor pairs, where \( f_i^k \) and \( \hat{g}_i^k \) are the descriptors from the original and reconstructed images, respectively. The descriptor loss is defined as:
\begin{equation}
\mathcal{L}_{\text{feat}} = \frac{1}{K} \sum_{i=1}^{K} \| f_i^k - \hat{g}_i^k \|_2^2.
\label{eq:feat}
\end{equation}
The total training loss combines the two objectives as:
\begin{equation}
\mathcal{L}_{\text{total}} = \lambda_1 \cdot \mathcal{L}_{\text{MSE}} + \lambda_2 \cdot \mathcal{L}_{\text{feat}},
\label{eq:total}
\end{equation}
where \( \lambda_1 \) and \( \lambda_2 \) control the trade-off between image reconstruction accuracy and descriptor-level consistency for 3D reconstruction.

\subsection{Pixel-wise Confidence Prediction Network}

To enhance the structural reliability of reconstructed images for downstream 3D reconstruction tasks, we introduce a pixel-wise confidence prediction network at the decoder side. The resulting confidence map estimates the reliability of each pixel in the reconstructed image and is used during subsequent camera pose estimation to identify structurally stable regions.

The confidence network takes the received semantic feature $\hat z_k \in \mathbb{R}^{\frac{H}{2^L} \times \frac{W}{2^L} \times 32C}$ as its initial input. This feature is first upsampled spatially and fused with the first-stage decoder feature. The fused feature is processed through a lightweight convolutional subnetwork. The output is then upsampled and fused with the next-stage decoder feature. This process repeats progressively across the decoder path, refining the confidence prediction at each level. The number of fusion stages is determined by the number of decoder layers in the backbone network. The final confidence map \( \mathbf {\hat c_k} \in \mathbb{R}^{H \times W} \) is activated through a Sigmoid function to constrain its values within the range \( [0,1] \).

During training, we construct a pseudo-label \(\mathbf  c_k^{\text{gt}} \) based on the absolute pixel-wise error between the reconstructed and original images:

\begin{equation}
\mathbf c_k^{\text{gt}}(i,j) = 1 - \tanh \left( \alpha \cdot \left| \hat{x}_k(i,j) - x_k(i,j) \right| \right).
\label{eq:conf_gt}
\end{equation}

Here, \( \alpha \) is a smoothing coefficient that adjusts the sensitivity to reconstruction error. A lower reconstruction error corresponds to a higher confidence value. The confidence network is trained using an L1 loss between the predicted map and the pseudo-label:

\begin{equation}
\mathcal{L}_{\text{conf}} = \frac{1}{N} \sum_{k=1}^{N} \| \mathbf {\hat c_k} - \mathbf c_k^{\text{gt}} \|_1.
\label{eq:conf_loss}
\end{equation}

The overall training objective combines three terms: a reconstruction loss, a feature consistency loss, and the confidence supervision loss:

\begin{equation}
\mathcal{L}_{\text{total}} = \lambda_1 \cdot \mathcal{L}_{\text{MSE}} + \lambda_2 \cdot \mathcal{L}_{\text{feat}} + \lambda_3 \cdot \mathcal{L}_{\text{conf}}.
\label{eq:total_loss}
\end{equation}

The weighting coefficients \( \lambda_1, \lambda_2, \lambda_3 \) control the trade-off between pixel-level accuracy, geometric consistency, and confidence map reliability during optimization. The training process is shown in Algorithm 1.

\begin{algorithm}[t]
\caption{Training Process of the 3DC-SC Semantic Transceiver}
\label{alg:3dc_sc}
\begin{algorithmic}[1]
    \State Initialization: Training image set $\mathcal{X}$, learning rate $\gamma$, 
    loss weights $(\lambda_1,\lambda_2,\lambda_3)$.     
    \Repeat
        \State \parbox[t]{\dimexpr\linewidth-\algorithmicindent}{
        Input: Training image $\mathbf{x}_{k} \in \mathbb{R}^{H \times W \times 1}$.}
        
        \State \parbox[t]{\dimexpr\linewidth-\algorithmicindent}{
        Semantic transmission through noisy channel:
        $\mathbf{x}_{k} \!\to\! 
        \mathcal{E}_e(\mathbf{x}_{k};\boldsymbol{\alpha})
        \!\to\! \mathbf{s}_{k}
        \!\to\! \hat{\mathbf{s}}_{k}
        \!\to\! \mathcal{E}_d(\hat{\mathbf{s}}_{k};\boldsymbol{\alpha})
        \!\to\! \hat{\mathbf{x}}_{k}$.}
        
        \State \parbox[t]{\dimexpr\linewidth-\algorithmicindent}{
        Pixel-wise confidence estimation:
        $\hat{\mathbf{x}}_{k} \to 
        \mathcal{E}_c(\hat{\mathbf{x}}_{k};\boldsymbol{\alpha})
        \to \hat{\mathbf{c}}_{k}$.}
        
        \State \parbox[t]{\dimexpr\linewidth-\algorithmicindent}{
        Compute reconstruction loss
        $\mathcal{L}_{\mathrm{MSE}} = 
        \|\mathbf{x}_{k}-\hat{\mathbf{x}}_{k}\|_2^2$.}
        
        \State \parbox[t]{\dimexpr\linewidth-\algorithmicindent}{
        Compute descriptor consistency loss $\mathcal{L}_{\mathrm{feat}}$.}
        
        \State \parbox[t]{\dimexpr\linewidth-\algorithmicindent}{
        Generate pseudo-label confidence map and compute confidence loss
        $\mathcal{L}_{\mathrm{conf}}$.}
        
        \State \parbox[t]{\dimexpr\linewidth-\algorithmicindent}{
        Update network parameters $\boldsymbol{\alpha}$ using total loss
        $\mathcal{L} = \lambda_1\mathcal{L}_{\mathrm{MSE}} +
        \lambda_2\mathcal{L}_{\mathrm{feat}} +
        \lambda_3\mathcal{L}_{\mathrm{conf}}$.}
    \Until{$t = T$}
\end{algorithmic}
\end{algorithm}

\section{Confidence-weighted Pose Estimation Algorithm Design}
In the 3D reconstruction stage, the receiver estimates the camera poses from the reconstructed image sequence $\hat{x}_k$ and their confidence maps $\hat{c}_k$.
For each consecutive image pair $(\hat{x}_k,\hat{x}_{k+1})$, the pose estimation module infers the relative camera motion represented by a rotation matrix $\mathbf{R}_{k,k+1}$ and a translation vector $\mathbf{T}_{k,k+1}$.
The overall procedure follows a feature–geometry based estimation framework: 
first, reliable keypoint correspondences are established between adjacent frames; 
then the relative pose $(\mathbf{R}_{k,k+1},\mathbf{T}_{k,k+1})$ is computed through robust geometric estimation under the epipolar constraint; 
finally, all poses are globally refined through a bundle adjustment (BA) optimization that jointly minimizes the reprojection residuals of multi-view correspondences.
During this process, the confidence maps $\hat{c}_k$ serve as pixel-wise reliability indicators that guide the selection and weighting of feature correspondences.
Specifically, the confidence information is integrated only into the RANSAC-based [24]\cite{ransac} geometric verification and the subsequent BA optimization, while all other components (including feature extraction, descriptor matching, and initial pose computation) follow the conventional processing pipeline used in widely adopted reconstruction frameworks such as ORB-SLAM2 and COLMAP\cite{colmap,orb}. This ensures compatibility with the common feature–geometry pipeline employed in mainstream SfM/SLAM tools.

\subsection{Feature Correspondence}
At the receiver side, the computing server estimates the relative camera poses based on the received reconstructed image sequence $\hat{x}_k$.  
Since geometric estimation relies on reliable point correspondences, we begin by describing how such correspondences are obtained.    

For two consecutive reconstructed frames $\hat{x}_k$ and $\hat{x}_{k+1}$, the rigid-body transformation $(\mathbf{R}_{k,k+1},\mathbf{T}_{k,k+1})$ between their camera coordinates satisfies the epipolar constraint:
\begin{equation}
(\mathbf{p}_i^{k+1})^{\top} \mathbf{F}\,\mathbf{p}_i^{k} = 0,\qquad 
\mathbf{F} = \mathbf{K}^{-T}[\mathbf{T}_{k,k+1}]_{\times}\mathbf{R}_{k,k+1}\mathbf{K}^{-1},
\label{eq:fundamental}
\end{equation}
where $\mathbf{K}$ is the intrinsic calibration matrix of the camera.

To obtain the feature correspondences, 
keypoint sets 
$\mathcal{P}_k=\{\mathbf{p}_i^{k}\}_{i=1}^{N_k}$ and 
$\mathcal{P}_{k+1}=\{\mathbf{p}_i^{k+1}\}_{i=1}^{N_{k+1}}$ 
are first extracted from the reconstructed images $\hat{x}_k$ and $\hat{x}_{k+1}$, respectively.
Local descriptors are then computed around each detected keypoint 
to characterize its visual features.
Each descriptor is represented as a numerical vector that summarizes 
the local image features of the corresponding keypoint.
Feature correspondences between 
$\mathcal{P}_k$ and $\mathcal{P}_{k+1}$ 
are established by matching these descriptors,
and the resulting correspondence set is denoted as 
$\mathcal{M}_k=\{(\mathbf{p}_i^{k},\mathbf{p}_i^{k+1})\}_{i=1}^{N}$.
In practice, $\mathcal{M}_k$ is obtained by performing nearest-neighbor search 
in the descriptor space as
\begin{equation}
\begin{aligned}
\mathcal{M}_k 
= \{(\mathbf{p}_i^{k},\mathbf{p}_i^{k+1}) \mid\ 
&\|d_i^{k} - d_i^{k+1}\|_2 
= \min_{j}\|d_i^{k} - d_j^{k+1}\|_2,\\
& i=1,\dots,N\}.
\end{aligned}
\label{eq:matches}
\end{equation}

Once the correspondence set is obtained, (\ref{eq:fundamental}) and the matched pairs jointly form the basis for solving the relative pose. However, due to reconstruction artifacts, texture ambiguity, and generative uncertainty, the correspondence set $\mathcal{M}_k$ typically contains a high proportion of mismatches. This significantly increases the difficulty of reliable geometric estimation. Therefore, an estimation mechanism is required to extract geometrically consistent correspondences. RANSAC is generally used as a traditional tool to solve this problem therefore propose a confidence-weighted RANSAC for fundamental matrix estimation \cite{ransac}.

\subsection{Confidence-weighted RANSAC}
Although RANSAC is a traditional tool for handling mismatched correspondences, its success depends on sampling an all-inlier minimal subset. A correspondence is considered an inlier if it is geometrically consistent with the underlying epipolar model; otherwise it is treated as an outlier. In our setting, reconstruction artifacts—such as blurred regions, structural distortions, or hallucinated details—introduce a large number of outliers, making uniform sampling unlikely to yield an all-inlier subset and thus reducing RANSAC stability. 
The confidence maps generated by the semantic transceiver provide pixel-level reliability cues that help distinguish trustworthy correspondences from uncertain ones. By integrating these confidence values into both the sampling and model evaluation stages, we obtain a more robust and efficient estimation procedure. We therefore propose a confidence-weighted RANSAC for fundamental matrix estimation \cite{ransac}.

Since each reconstructed frame $\hat{x}_k$ is accompanied by a confidence map $\mathbf{\hat{c}}_k$, the confidence of each correspondence $(\mathbf{p}_i^{k},\mathbf{p}_i^{k+1})$ is defined by fusing the two per-frame pixel confidences sampled at the matched keypoint locations:
\begin{equation}
c_i = \min\big(\mathbf{\hat{c}}_k(\mathbf{p}_i^{k}),\,
              \mathbf{\hat{c}}_{k+1}(\mathbf{p}_i^{k+1})\big),
\label{eq:confidence_fusion_revised}
\end{equation}
where $\mathbf{\hat{c}}_k(\mathbf{p}_i^{k})$ denotes the scalar confidence value at the 2D coordinates of $\mathbf{p}_i^{k}$.  
The fused confidence $c_i \in [0,1]$ measures the joint reliability of the matched keypoint pair.

Given the matching set $\mathcal{M}_k=\{(\mathbf{p}_i^{k},\mathbf{p}_i^{k+1})\}_{i=1}^{N}$, 
the estimation proceeds by repeatedly drawing a minimal subset of size $m$ from $\mathcal{M}_k$ to instantiate a candidate fundamental matrix $\mathbf{F}_t$ in the $t$-th trial.  
A minimal subset contains the smallest number of correspondences required to solve for the fundamental matrix (e.g., five points for the five-point algorithm or eight points for the eight-point algorithm \cite{ransac}), so that each sampled subset yields a complete hypothesis of the underlying epipolar geometry. 

To make the subset selection reflect the reliability of observations, 
a weighted sampling without replacement is adopted: 
the sampling distribution over correspondences is defined as
\begin{equation}
q_i=\frac{c_i}{\sum_{j=1}^{N} c_j},
\label{eq:ran_sampling}
\end{equation}
so that matches with higher confidence are more likely to be included in the minimal subset.  
Each candidate $\mathbf{F}_t$ is estimated from the selected subset and can subsequently be decomposed into an initial relative pose $(\mathbf{R}_{k,k+1}^{(0)}, \mathbf{T}_{k,k+1}^{(0)})$ according to Eq.~(\ref{eq:fundamental}).
For the estimated $\mathbf{F}_t$, each correspondence $(\mathbf{p}_i^{k},\mathbf{p}_i^{k+1})$ is evaluated by its geometric residual $r_i$, 
which measures the deviation from the epipolar constraint in (\ref{eq:fundamental}):
\begin{equation}
r_i=\big|\,(\mathbf{p}_i^{k+1})^{\top}\mathbf{F}_t\mathbf{p}_i^{k}\,\big|.
\label{eq:ran_residual}
\end{equation}
A correspondence is considered geometrically consistent if its residual is below a confidence-adaptive threshold,
\begin{equation}
\delta_i=\frac{\delta_0}{\sqrt{c_i+\varepsilon}},
\label{eq:ran_threshold}
\end{equation}
where $\delta_0$ is the base threshold and $\varepsilon$ is a small constant to avoid division by zero. 
After evaluating all candidates $\mathbf{F}_t$ over $T$ iterations, 
the algorithm follows the standard RANSAC procedure and selects the model that yields the largest number of inliers:
\begin{equation}
\mathbf{F}^\star = \arg\max_{\mathbf{F}_t} \ |\mathcal{I}(\mathbf{F}_t)|,
\end{equation}
where $\mathcal{I}(\mathbf{F}_t)=\{\, i \mid r_i < \delta_i \,\}$ denotes the set of correspondences satisfying the confidence-adaptive geometric residual constraint. 

\textbf{Remark:}
When the confidence values are positively correlated with the true inlier probabilities, biasing the RANSAC sampling distribution toward high-confidence correspondences increases the expected inlier ratio of a sampled minimal subset compared with uniform sampling.
However, this does not justify deterministically selecting the top-$m$ correspondences
based solely on confidence.
Deterministic top-$m$ selection fixes the minimal subset across all iterations,
forcing the estimation to rely on a single, potentially biased set of matches.
If these correspondences are spatially clustered
(e.g., lying on a single plane or object boundary),
or if even one incorrect match is assigned an overly high confidence,
the resulting subset becomes geometrically degenerate
or consistently leads to an incorrect model.
In such cases, RANSAC cannot recover, since no alternative subsets are explored.
The proposed confidence-weighted sampling preserves stochastic exploration while favoring reliable correspondences.

\subsection{Confidence-weighted BA Optimization}
The refined fundamental matrix $\mathbf{F}^\star$ estimated in the previous stage is decomposed into the initial relative pose $(\mathbf{R}_{k,k+1}^{(0)}, \mathbf{T}_{k,k+1}^{(0)})$ according to the epipolar formulation in Eq.~(\ref{eq:fundamental}). Then, a global multi-view optimization is performed to further refine the overall camera trajectory and the reconstructed 3D structure. 
This process is typically formulated as BA (BA), 
which jointly optimizes all camera poses and 3D points by minimizing the reprojection error 
between the observed image points and the projected positions of the corresponding 3D points 
in each image. 
Conventional BA assumes that all observations are equally reliable, 
which may not hold in our scenario due to transmission noise, reconstruction artifacts, 
and uncertainty introduced by generative reconstruction. 
As a result, directly treating all observations with equal weight may degrade 
the accuracy and robustness of global pose refinement.

To address this issue, we incorporate the confidence information into the BA framework.
such that observations with higher confidence exert stronger influence in the optimization, 
while those with lower confidence are adaptively down-weighted. 
The confidence of $\mathbf{p}_i^k$ in the $k$-th image is denoted as
\begin{equation}
c_i^k = \mathbf{\hat{c}}_k(\mathbf{p}_i^k).
\end{equation}

We modify the conventional BA objective used in COLMAP\cite{colmap} and ORB-SLAM2\cite{orb} by introducing pixel-wise confidence as multiplicative observation weights. This yields the following confidence-weighted formulation:
\begin{equation}
\min_{\{\mathbf{R}_k,\mathbf{T}_k\},\,\mathcal{P}}
\sum_{k} \sum_{i\in\mathcal{V}_k}
w_i^k\,\rho\!\left(
\left\|
\mathbf{p}_i^k - 
\pi\!\big(\mathbf{K}[\mathbf{R}_k\mid \mathbf{T}_k]\mathbf{P}_i\big)
\right\|_2^{2}
\right),
\label{eq:ba_objective}
\end{equation}
where $\mathcal{V}_k$ denotes the set of 3D points visible in the $k$-th image, 
$\pi(\cdot)$ is the projection function, $w_i^k$ is the observation weight
and $\rho(\cdot)$ is a robust loss function.  

The pixel-wise confidence produced by the semantic decoder is treated as an observation reliability indicator and is directly mapped to the reprojection noise variance through a monotonic decreasing function. The observation weight $w_i^k$ is determined by the confidence value as
\begin{equation}
w_i^k = (\sigma_i^k)^{-2}, \qquad
(\sigma_i^k)^2 = \sigma_0^2 \exp\!\big(\beta(1 - c_i^k)\big),
\label{eq:ba_weight}
\end{equation}
where $\sigma_0^2$ is the base noise variance and $\beta$ is a scaling coefficient.  
A higher confidence $c_i^k$ leads to a smaller variance $(\sigma_i^k)^2$ 
and thus a larger weight $w_i^k$, 
providing stronger geometric constraints during optimization, 
while low-confidence observations are automatically down-weighted to enhance robustness.

The confidence-weighted BA objective in (\ref{eq:ba_objective}) is a nonlinear optimization problem 
due to the camera projection function. 
It is therefore solved iteratively using the Gauss–Newton algorithm \cite{colmap}. 
At each iteration, the reprojection residuals are linearized around the current estimate 
to compute the parameter increment. 
Let $\boldsymbol{\xi}$ denote the incremental update of all optimization variables 
(including camera poses and 3D points), 
and let $\mathbf{r}$ be the stacked vector of all reprojection residuals. 
A first-order Taylor expansion yields the linear approximation
\begin{equation}
\mathbf{r} \;\approx\; \mathbf{J}\boldsymbol{\xi} + \boldsymbol{\epsilon},
\label{eq:ba_linear}
\end{equation}
where $\mathbf{J}$ is the Jacobian matrix of the residuals with respect to the parameters, 
and $\boldsymbol{\epsilon}$ denotes the observation noise.  
According to the confidence-dependent variance model in (\ref{eq:ba_weight}), 
the noise covariance takes the form
\begin{equation}
\mathrm{Cov}(\boldsymbol{\epsilon})
= \boldsymbol{\Sigma}
= \mathrm{diag}\!\left((\sigma_i^k)^2 \mathbf{I}_2\right).
\end{equation}

Under this linearized model, the parameter increment $\boldsymbol{\xi}$ is obtained 
by solving a confidence-weighted least-squares problem, 
whose closed-form solution is
\begin{equation}
\hat{\boldsymbol{\xi}}
= (\mathbf{J}^\top \mathbf{W}\mathbf{J})^{-1}
  \mathbf{J}^\top \mathbf{W}\mathbf{r},
\qquad
\mathbf{W} = \mathrm{diag}(w_i^k \mathbf{I}_2).
\label{eq:ba_solution}
\end{equation}

\begin{algorithm}[t]
    \caption{Confidence-weighted Pose Estimation and 3D Reconstruction}
    \label{alg:conf_pose_ba}
    \begin{algorithmic}[1]
        \State Initialization: reconstructed images $\{\mathbf{\hat{x}}_{k}\}_{k=1}^{n}$, confidence maps $\{\mathbf{\hat{c}}_{k}\}_{k=1}^{n}$, intrinsics $\mathbf{K}$.
        
        \For{$k = 1$ to $n-1$}
            \State \parbox[t]{\dimexpr\linewidth-\algorithmicindent}{
            Extract keypoints and descriptors on $(\mathbf{\hat{x}}_{k},\mathbf{\hat{x}}_{k+1})$, 
            match descriptors to obtain $\mathcal{M}_k=\{(\mathbf{p}_i^{k},\mathbf{p}_i^{k+1})\}_{i=1}^{N}$.}
            
            \State \parbox[t]{\dimexpr\linewidth-\algorithmicindent}{
            For each match $(\mathbf{p}_i^{k},\mathbf{p}_i^{k+1})$, fuse pixel-wise confidences
            $c_i = \min(\mathbf{\hat{c}}_k(\mathbf{p}_i^{k}),\mathbf{\hat{c}}_{k+1}(\mathbf{p}_i^{k+1}))$,
            define sampling probability $q_i \propto c_i$, 
            and run confidence-weighted RANSAC to estimate
            initial relative pose $(\mathbf{R}_{k,k+1}^{(0)},\mathbf{T}_{k,k+1}^{(0)})$.}
        \EndFor
        
        \State \parbox[t]{\dimexpr\linewidth-\algorithmicindent}{
        Initialize global poses $\{\mathbf{R}_k,\mathbf{T}_k\}_{k=1}^{n}$ and sparse points $\mathcal{P}$ 
        from relative poses and multi-view correspondences.}
        
        \Repeat
            \State \parbox[t]{\dimexpr\linewidth-\algorithmicindent}{
            For each observation of $\mathbf{P}_i$ at $\mathbf{p}_i^k$, query confidence 
            $c_i^k = \mathbf{\hat{c}}_k(\mathbf{p}_i^k)$, 
            set variance $(\sigma_i^k)^2 = \sigma_0^2 \exp(\beta(1-c_i^k))$ 
            and weight $w_i^k = (\sigma_i^k)^{-2}$, 
            and construct diagonal matrix $\mathbf{W} = \mathrm{diag}(w_i^k\mathbf{I}_2)$.}
            
            \State \parbox[t]{\dimexpr\linewidth-\algorithmicindent}{
            Form stacked reprojection residual vector $\mathbf{r}$ and Jacobian $\mathbf{J}$,
            solve the confidence-weighted normal equations
            $\hat{\boldsymbol{\xi}} = (\mathbf{J}^\top \mathbf{W}\mathbf{J})^{-1}\mathbf{J}^\top \mathbf{W}\mathbf{r}$,
            and update $\{\mathbf{R}_k,\mathbf{T}_k\}$ and $\mathcal{P}$ accordingly.}
        \Until{BA converges}
        
        \State Output: refined camera poses $\{\mathbf{R}_k,\mathbf{T}_k\}_{k=1}^{n}$ and sparse 3D point set $\mathcal{P}$.
    \end{algorithmic}
\end{algorithm}
The estimated increment $\hat{\boldsymbol{\xi}}$ is then used to update the camera poses and 3D points, and the procedure is repeated until convergence. 
This linearized formulation reveals that each BA iteration solves a weighted linear system, where the confidence values directly modulate the influence of individual observations. To characterize the statistical effect of the confidence-weighted update in the linearized BA system, we provide the following proposition.

\textbf{Proposition~1.}
Under the linearized model in Eq.~(\ref{eq:ba_linear}),
the estimator in Eq.~(\ref{eq:ba_solution}) with $\mathbf{W}=\boldsymbol{\Sigma}^{-1}$ 
is the best linear unbiased estimator (BLUE).
Its covariance matrix is given by
\begin{equation}
\mathrm{Cov}(\hat{\boldsymbol{\xi}}) = (\mathbf{J}^\top \mathbf{W}\mathbf{J})^{-1}.
\end{equation}

\textit{Proof.}
Under the linearized model 
$\mathbf{r}=\mathbf{J}\boldsymbol{\xi}+\boldsymbol{\epsilon}$ 
with zero-mean noise and covariance 
$\mathrm{Cov}(\boldsymbol{\epsilon})=\boldsymbol{\Sigma}$,
the Gauss--Markov theorem states that the linear unbiased estimator 
with minimum variance is obtained by solving the weighted least-squares problem 
with weight matrix $\mathbf{W}=\boldsymbol{\Sigma}^{-1}$. 
Substituting $\mathbf{W}=\boldsymbol{\Sigma}^{-1}$ into Eq.~(\ref{eq:ba_solution}) 
gives
\[
\hat{\boldsymbol{\xi}}
= (\mathbf{J}^\top \boldsymbol{\Sigma}^{-1}\mathbf{J})^{-1}
  \mathbf{J}^\top \boldsymbol{\Sigma}^{-1}\mathbf{r},
\]
which is therefore the best linear unbiased estimator (BLUE). 
Its covariance directly follows as 
$\mathrm{Cov}(\hat{\boldsymbol{\xi}}) = (\mathbf{J}^\top \mathbf{W}\mathbf{J})^{-1}$. 
\hfill$\square$

This proposition shows that the confidence-weighted BA provides,
at each linearization step, the minimum-variance linear unbiased estimate of the pose and structure parameters.
By incorporating pixel-wise confidence as heteroscedastic observation weights,
the optimization adaptively emphasizes reliable observations and reduces uncertainty introduced by low-confidence regions. The processing of the proposed semantic-aware pose estimation algorithm is described in Algorithm 2.

\begin{figure*}[htp]
	\centering
        \setlength{\abovecaptionskip}{-0pt}
        \setlength{\belowcaptionskip}{-0pt}
	\includegraphics[width=0.9\textwidth]{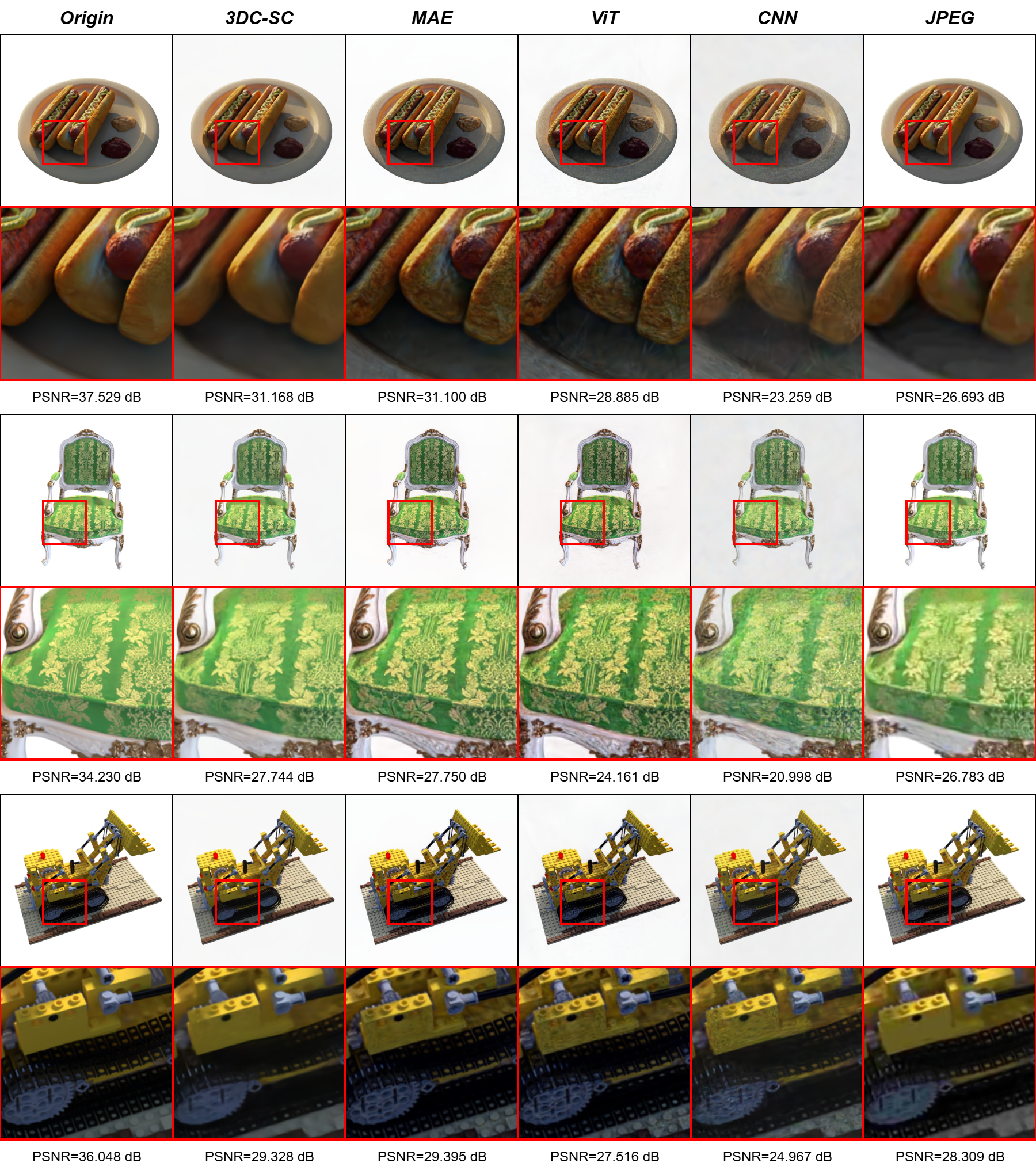}
  	\caption{3DGS reconstruction results of 3DC-SC and other benchmarks under AWGN channel with SNR=10 dB}
	\label{img2}
\end{figure*}

\section{Simulation Results}
\subsection{Simulation Setup}
In this section, we conduct simulations to evaluate the performance of the proposed 3DC-SC in terms of image reconstruction quality, pose estimation accuracy, and overall 3D reconstruction quality. Four benchmark schemes, including MAE, ViT, CNN, and JPEG+LDPC, are considered for comparison to validate the advantages of SemCom under noisy channels. All methods are tested under identical channel conditions and reconstruction methods to ensure fairness.

For training the semantic transceivers, we use the ImageNet dataset containing 14,197,122 images, enabling all models to learn robust image reconstruction. For 3D reconstruction simulation, we adopt Blender-rendered multi-view sequences consisting of eight 3D object models, each rendered from 100 viewpoints with an image resolution of \(800 \times 800\). This dataset serves as the evaluation benchmark for analyzing the impact of communication noise on geometric consistency. For image reconstruction quality, we use the Kodak24 as the test set, which includes 24 different \(768 \times 512\) images.

The channel model is an additive white Gaussian noise (AWGN) channel with 16-QAM modulation. The signal-to-noise ratio (SNR) ranges from 2 dB to 12 dB to assess robustness under different noise conditions. The proposed 3DC-SC have 4 downsample block with a learning rate of \(10^{-4}\), a batch size of 16 and the Adam optimizer. All models are implemented in PyTorch and trained on an NVIDIA RTX A6000 GPU.

The benchmark methods considered in this study are as follows:
\begin{itemize}
\item \textit{MAE:}  
A semantic coding model based on the MAE architecture. For the \(800 \times 800\) images used in 3D reconstruction, the number of patches is \(50 \times 50\). 

\item \textit{ViT:}  
A semantic transceiver built upon a Vision Transformer encoder--decoder. It represents a typical transformer-based semantic coding scheme.

\item \textit{CNN:}  
A representative benchmark for traditional depth image encoders.

\item \textit{JPEG+LDPC:}  
A classical separated source--channel coding pipeline, where JPEG is used for image compression and LDPC with rate \(1/2\) is used for channel coding. This scheme does not provide semantic-level representations and serves as a non-learning traditional communication benchmark.
\end{itemize}

To ensure fairness, all neural benchmarks are trained using the same training dataset, optimizer configuration, and channel conditions. Moreover, all reconstructed image sequences undergo an identical 3D reconstruction procedure, ensuring that any performance differences arise solely from the semantic encoding structures and decoding strategies of the corresponding models.

\subsection{Results Analysis}
\subsubsection{3D reconstruction Performance}
In this part, we compare the 3D reconstruction results obtained under an SNR of 10 dB. 
For each SemCom method, the transmitted images are used to reconstruct a 3DGS model, and a fixed-view rendering of the reconstructed scene is presented \cite{3DGS}. 
We consider Hotdog, Chair and Lego scenes, visualize key local regions to examine differences in texture fidelity, high-frequency details, and overall visual consistency.

Fig. 3 presents the visual results of 3D reconstruction, 3DC-SC produces natural and continuous textures without noticeable artifacts, and its structural clarity is well preserved. 
The overall PSNR of 3DC-SC is close to that of MAE, despite being a significantly lighter model. 
MAE achieves the sharpest structural boundaries, but visible artifacts appear in several high-frequency regions. 
ViT exhibits blurred or fragmented details in complex textured areas, suggesting higher sensitivity to channel noise. 
CNN produces overly smoothed reconstructions with substantial loss of high-frequency information. 
JPEG performs reasonably well at this SNR level, with moderate texture degradation but without extreme distortions, and maintains a mid-range PSNR.
These differences arise from the inherent architectural characteristics and noise robustness of each method. 
3DC-SC employs a lightweight convolutional structure and is trained end-to-end, enabling stable feature representations that preserve natural textures and avoid artifact formation. 
MAE, despite its superior structural recovery, relies on mask-based reconstruction, which tends to introduce compensatory artifacts. Moreover, its model size is significantly larger than 3DC-SC, making it impractical for deployment on resource-constrained devices. 
Achieving reconstruction quality close to MAE with a much smaller model is therefore a key advantage of 3DC-SC. 
JPEG performs adequately at SNR = 10 dB because the channel condition is not severely degraded and the JPEG+LDPC pipeline can still effectively suppress noise. 
However, its limitations become pronounced under lower SNR levels, which will be further demonstrated in the subsequent line-plot analysis.

\begin{figure}[htbp]
    \centering
        \setlength{\abovecaptionskip}{-0pt}
        \setlength{\belowcaptionskip}{-0pt}
        \vspace{-0cm}
    \includegraphics[width=0.45\textwidth]{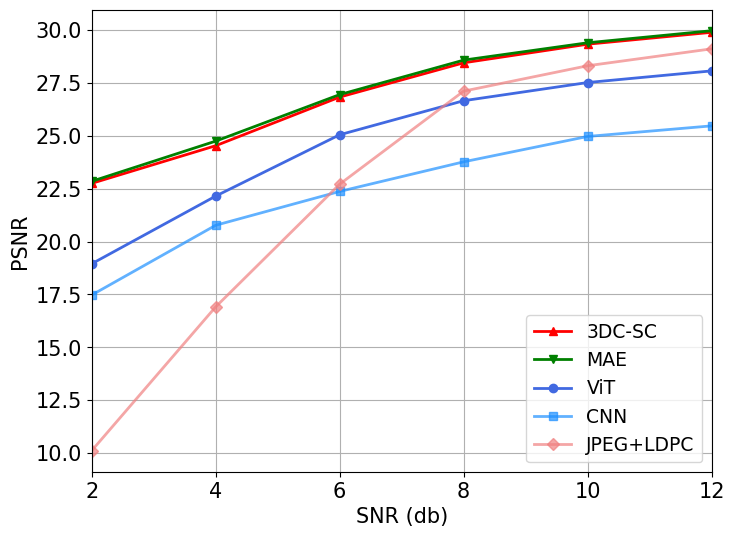}
    \caption{3DGS reconstruction results average PSNR of 3DC-SC and other benchmarks under AWGN channel under different SNR}
    \label{img2}
\end{figure}
\begin{figure}[htbp]
    \centering
        \setlength{\abovecaptionskip}{-0pt}
        \setlength{\belowcaptionskip}{-0pt}
        \vspace{-0cm}
    \includegraphics[width=0.45\textwidth]{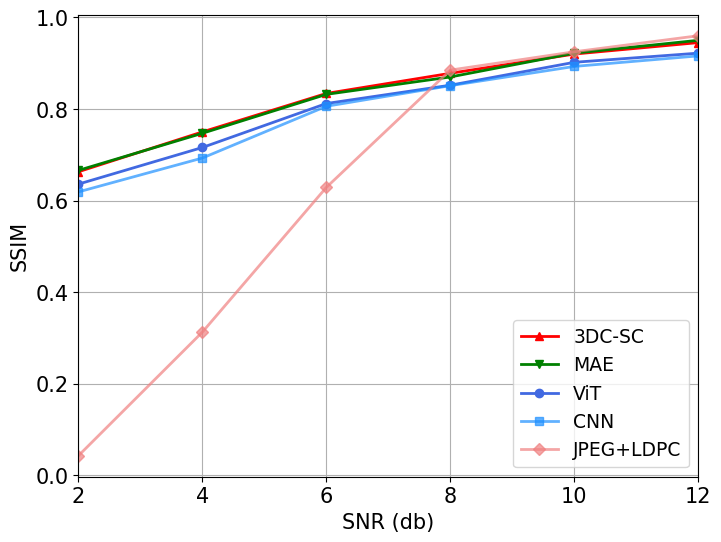}
    \caption{3DGS reconstruction results average SSIM of 3DC-SC and other benchmarks under AWGN channel under different SNR}
    \label{img2}
\end{figure}

Fig. 4 and Fig. 5 present the PSNR and SSIM curves of different methods as the SNR varies from 2 dB to 12 dB, reflecting the robustness of each approach in preserving the quality of 3DGS-rendered images under different channel conditions. 
PSNR characterizes the pixel-wise intensity deviation, while SSIM captures the structural similarity between reconstructed and reference renderings.
Both figures show that 3DC-SC and MAE consistently achieve the highest performance across the entire SNR range, with their trajectories remaining very close to each other. 
ViT and CNN follow behind, with noticeably lower values at low SNR levels, whereas JPEG+LDPC experiences severe degradation in low-SNR conditions but rises sharply as the SNR increases, approaching some neural baselines at higher SNR values. 
These trends highlight the fundamental advantages of SemCom over traditional JPEG coding. 
SemCom models learn noise-robust feature representations, enabling smooth and monotonic quality improvement as the SNR increases, and maintaining high reconstruction fidelity even under challenging channel conditions. 
In contrast, JPEG+LDPC, as a non-semantic pixel-level coding scheme, is highly sensitive to bit errors, resulting in sharp performance drops at low SNR and recovery only at higher SNR where LDPC error correction becomes effective. 
\begin{figure*}[htp]
	\centering
        \setlength{\abovecaptionskip}{-0pt}
        \setlength{\belowcaptionskip}{-0pt}
	\includegraphics[width=0.9\textwidth]{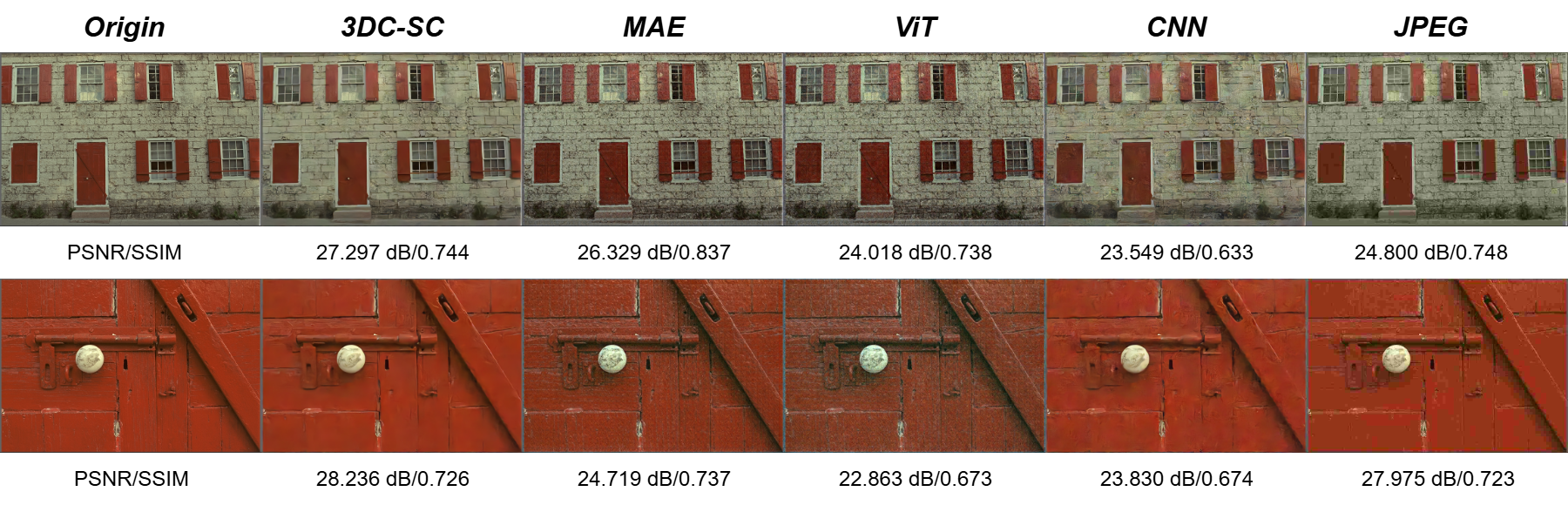}
  	\caption{Comparison of 3DC-SC and other benchmarks reconstructed images under AWGN channel with SNR=10 dB}
	\label{img2}
\end{figure*}
\begin{figure}[htbp]
    \centering
        \setlength{\abovecaptionskip}{-0pt}
        \setlength{\belowcaptionskip}{-0pt}
        \vspace{-0cm}
    \includegraphics[width=0.45\textwidth]{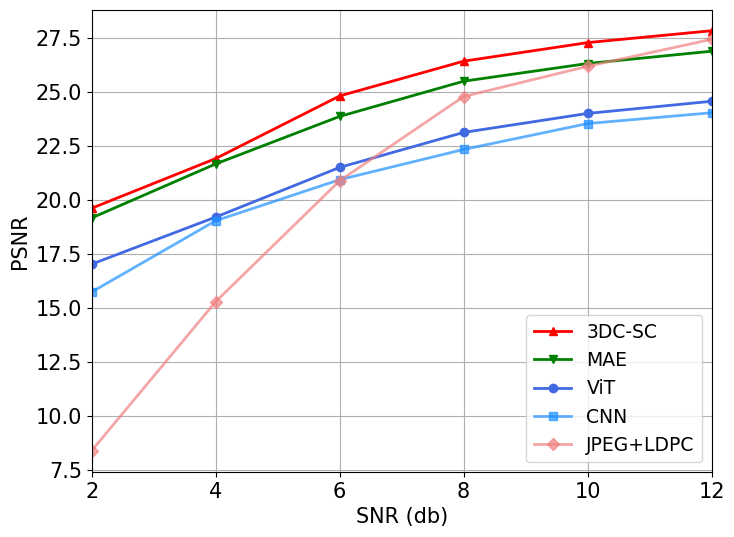}
    \caption{Comparison of 3DC-SC and other benchmarks reconstructed images PSNR under AWGN channel under different SNR}
    \label{img2}
\end{figure}
\begin{figure}[htbp]
    \centering
        \setlength{\abovecaptionskip}{-0pt}
        \setlength{\belowcaptionskip}{-0pt}
        \vspace{-0cm}
    \includegraphics[width=0.45\textwidth]{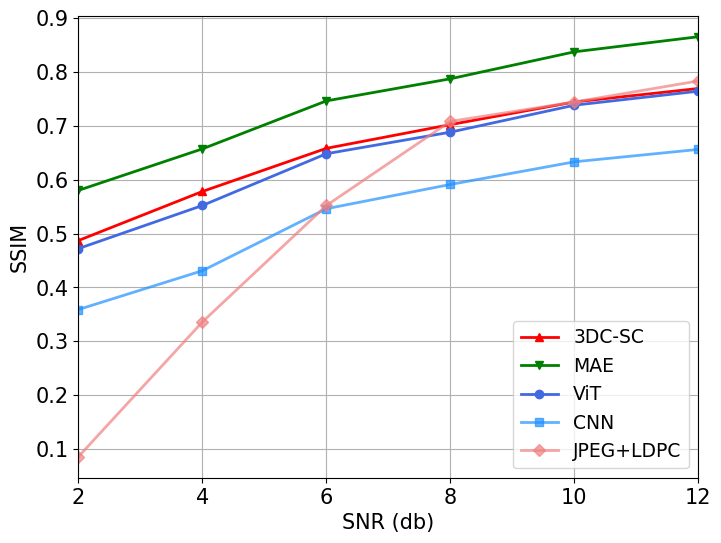}
    \caption{Comparison of 3DC-SC and other benchmarks reconstructed images SSIM under AWGN channel under different SNR}
    \label{img2}
\end{figure}

\subsubsection{Image reconstructed Performance}
Fig.~6 shows the reconstruction results of different methods on two images from the Kodak24 dataset under an SNR of 10 dB, together with the corresponding PSNR and SSIM values.
In these comparisons, 3DC-SC produces natural and consistent textures in both images, with stable color reproduction and no noticeable artifacts, achieving relatively high PSNR and SSIM scores. MAE preserves structural details particularly well, especially around building contours and window or door edges, although certain local regions exhibit visible reconstruction artifacts. ViT yields overall balanced results, but high-frequency details tend to appear slightly blurred, while CNN produces overly smoothed outputs with substantial loss of fine textures. JPEG maintains acceptable global quality at this SNR level, yet pronounced block artifacts and color banding remain visible in regions with complex textures.
These reconstruction results align with the trends observed earlier in the 3D reconstruction simulations . The difference is that, in 2D image reconstruction, the reconstruction discrepancy mainly manifests as variations in texture fidelity, smoothness, and artifact types. In contrast, within 3D reconstruction pipelines, the same discrepancies further amplify into differences in structural consistency, view stability, and overall geometric reconstruction accuracy. Although the visual differences between 2D and 3D settings are not identical, the results remain consistent with our earlier findings. Models that preserve clean, artifact-free textures provide more reliable inputs for geometric reasoning. Notably, the performance of 3DC-SC in 3D reconstruction is not merely a consequence of higher 2D fidelity; its ability to maintain texture continuity and local structural coherence is particularly aligned with the needs of multi-view geometry. Thus, the improvements observed in 3D tasks stem from their confidence-guide design rather than from image quality alone.

Fig.~7 and Fig.~8 presents the PSNR and SSIM curves of different methods on the Kodak24 dataset as the SNR varies from 2 dB to 12 dB. 
Across both metrics, all methods exhibit monotonically increasing performance as the SNR improves. 
In the PSNR plot, 3DC-SC consistently achieves the highest values over the entire SNR range, closely followed by MAE, while ViT and CNN remain lower. 
JPEG+LDPC suffers severe degradation at low SNR but recovers sharply as the channel condition improves, approaching some neural baselines in the high-SNR region. 
The SSIM curve follows a similar rising trend; however, MAE attains the highest SSIM at all SNR levels, with 3DC-SC ranking second. 
ViT and CNN exhibit noticeably weaker structural preservation, especially under low SNR, where the performance gap becomes more pronounced.
These results demonstrate the robustness advantage of SemCom methods over traditional JPEG-based coding. 
3DC-SC outperforms all baselines in PSNR while maintaining competitive SSIM, achieving performance close to or exceeding that of MAE despite using a significantly lighter model. 
JPEG+LDPC, as a non-semantic scheme, remains highly sensitive to channel noise and performs poorly under low SNR, only recovering to acceptable quality when strong error correction compensates at high SNR. 
\setcounter{figure}{10}
\begin{figure*}[htp]
	\centering
        \setlength{\abovecaptionskip}{-0pt}
        \setlength{\belowcaptionskip}{-0pt}
	\includegraphics[width=0.9\textwidth]{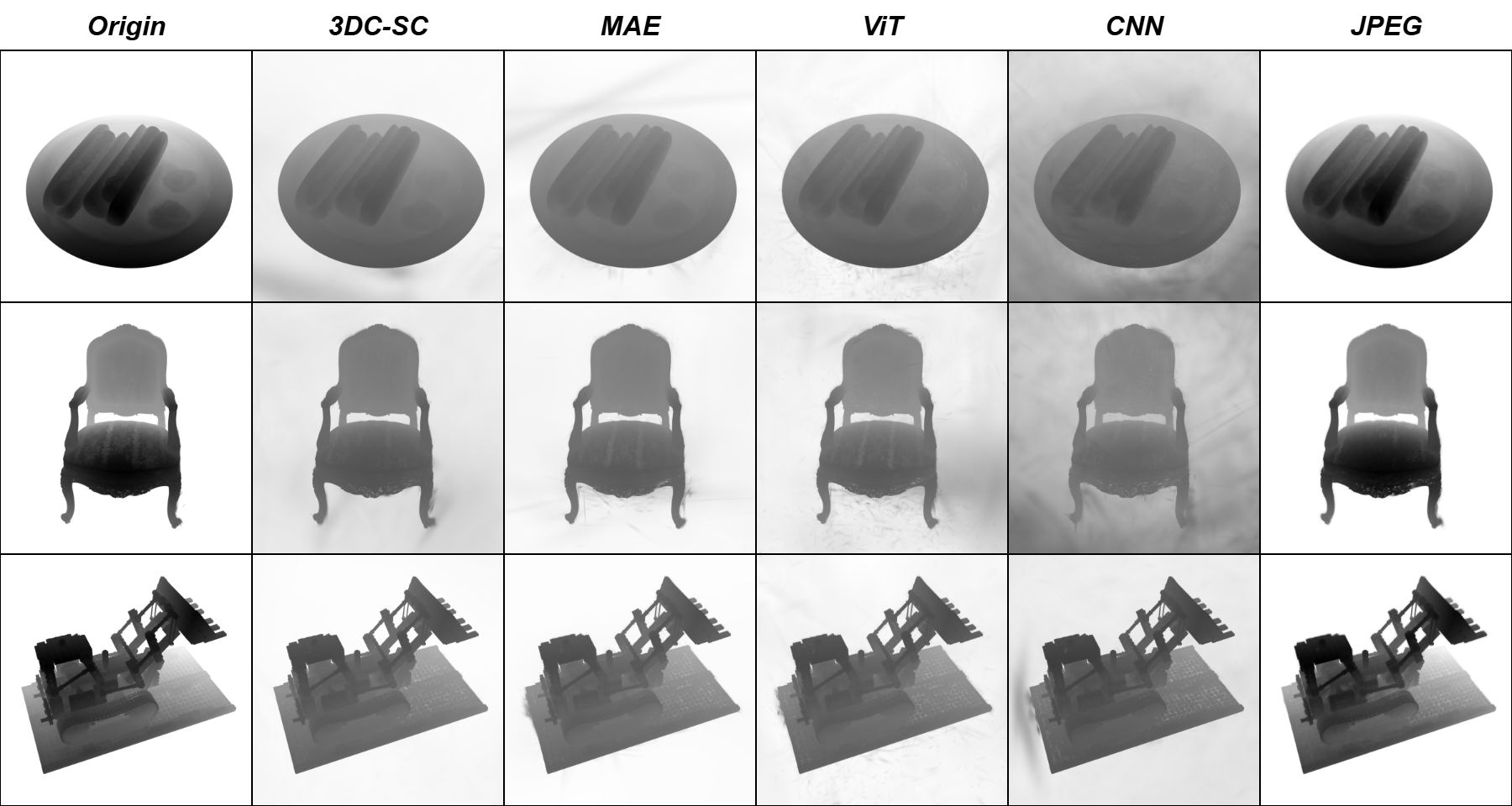}
  	\caption{Comparison of the depth map of 3DC-SC and other benchmarks reconstructed images under AWGN channel with SNR=10 dB}
	\label{img2}
\end{figure*}
\setcounter{figure}{8}
\begin{figure}[htbp]
    \centering
        \setlength{\abovecaptionskip}{-0pt}
        \setlength{\belowcaptionskip}{-0pt}
        \vspace{-0cm}
    \includegraphics[width=0.45\textwidth]{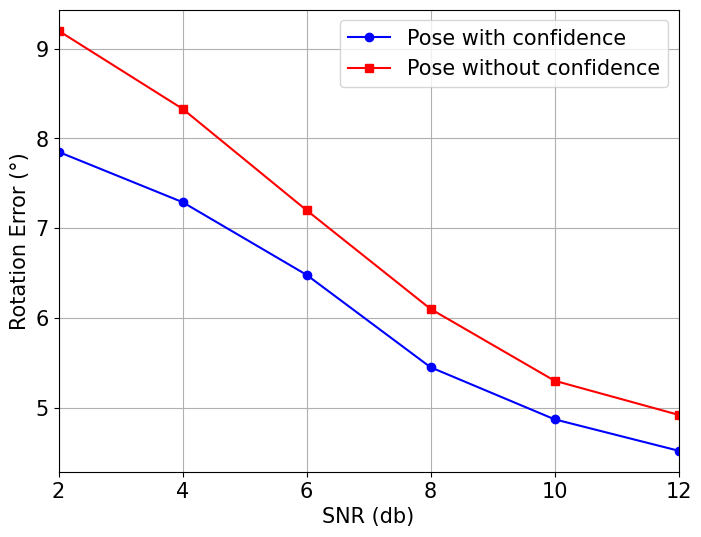}
    \caption{Rotation error of pose estimation with confidence and pose estimation without confidence compared to the original image under AWGN channel under different SNR}
    \label{img2}
\end{figure}
\begin{figure}[htbp]
    \centering
        \setlength{\abovecaptionskip}{-0pt}
        \setlength{\belowcaptionskip}{-0pt}
        \vspace{-0cm}
    \includegraphics[width=0.47\textwidth]{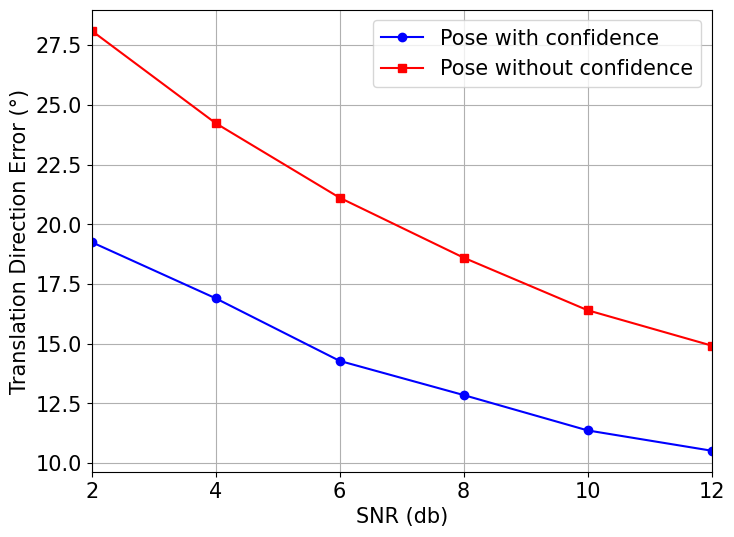}
    \caption{Translation direction error of pose estimation with confidence and pose estimation without confidence compared to the original image under AWGN channel under different SNR}
    \label{img2}
\end{figure}

\subsubsection{Pose Estimation Performance}
Fig.~9 and Fig.~10 reports the average rotation error and translation-direction error obtained on the Blender dataset under SNR levels ranging from 2~dB to 12~dB. 
For each SNR setting, all multi-view images of the Blender scenes are transmitted through our 3DC-SC transceiver either with confidence or without confidence and the resulting reconstructed images are used to estimate camera poses. 
The reported curves therefore reflect the averaged pose estimation accuracy across all viewpoints.

Both rotation and translation-direction errors decrease significantly as the SNR increases, indicating that higher-quality reconstructed images lead to more stable geometric inference. 
Across all SNR levels, however, the confidence-enabled estimator consistently outperforms the estimator without confidence. 
For rotation, the improvement is around 1–2$^\circ$ throughout the SNR range, whereas for translation-direction the advantage is even more pronounced, reaching 8–10$^\circ$ under low-SNR conditions (2–6~dB). 
Although the performance gap narrows as the SNR rises, the confidence-enabled estimator maintains the lowest error in all cases.

This performance gain arises from the pixel-wise reliability information provided by the confidence map. 
By highlighting regions with higher reconstruction fidelity, the confidence allows the system to emphasize reliable features during matching, RANSAC inlier selection, and subsequent BA, while suppressing the influence of noised areas. 
Consequently, the confidence-guided pipeline yields more robust and accurate pose estimates, particularly under noisy channel conditions, whereas the confidence-free approach is more vulnerable to degradation.

\subsubsection{Depth Map Performance}

Fig.~11 presents the depth maps rendered from the 3DGS reconstruction under an SNR of 10~dB for three representative Blender scenes. 
An interesting observation is that, under this moderately noisy channel condition, the depth maps obtained using JPEG+LDPC are almost indistinguishable from the ground truth and in some cases visually outperform those produced by several SemCom methods. This differs from the trends observed in appearance and texture reconstruction. The reason lies in the fact that depth maps reflect a different aspect of the 3D reconstruction pipeline and respond differently to distortion types. In our setting, JPEG introduces primarily smooth and locally coherent compression artifacts, and 3DGS being optimized through multi-view photometric consistency—tends to be robust to such distortions. In contrast, SemCom methods may introduce slight generative biases or multi-view inconsistencies that are negligible in 2D appearance but can accumulate during multi-view optimization, leading to small deviations in depth rendering.
However, this does not suggest that traditional methods hold an overall advantage in 3D reconstruction. SemCom remains markedly more robust in low-SNR regimes and provides more stable performance on tasks that depend strongly on semantic or structural cues. The strength of JPEG at SNR = 10~dB simply highlights that different 3D reconstruction outputs have distinct sensitivities to distortion. This observation underscores that depth-focused or geometry-oriented applications may require dedicated semantic feature designs tailored to their characteristics, rather than relying solely on objectives optimized for appearance reconstruction.

\section{Conclusion}
In this paper, we propose a SemCom framework for real-time mobile 3D reconstruction, which focuses on transmitting geometry-relevant semantic information rather than pixel-level fidelity to support robust multi-view estimation under imperfect communication conditions. By jointly delivering reconstructed images and pixel-wise confidence maps, the proposed framework enables downstream geometric modules to explicitly account for communication-induced uncertainty.
Guided by this design, we develop a semantic transceiver and confidence-guided geometric estimation scheme that integrate semantic reliability into pose estimation and multi-view optimization. Simulation results demonstrate that the proposed framework improves the robustness and accuracy of camera pose estimation and 3D reconstruction while maintaining efficient communication and lightweight model complexity.
This work provides insights into the integration of SemCom and geometry-sensitive vision tasks by emphasizing that communication reliability should be evaluated through its impact on multi-view geometric estimation, rather than appearance fidelity alone. It highlights the importance of aligning SemCom with the requirements of real-time and incremental 3D reconstruction, thereby moving beyond conventional appearance-oriented transmission strategies.

\bibliographystyle{IEEEtran}
\bibliography{IEEEexample}
\vfill

%

\begin{IEEEbiography}
[{\includegraphics[width=1in,height=1.25in,clip,keepaspectratio]{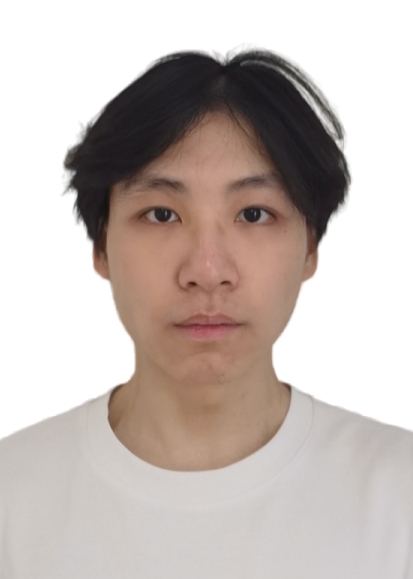}}]
{Fangzhou Zhao} received the M.Sc. degree and Ph.D degree from the James Watt School of Engineering, University of
Glasgow, Glasgow, U.K., in 2020 and 2025, respectively. He is currently a Lecturer and also a Postdoctoral Researcher with North China Electric Power University. His research interests include deep learning in wireless communication and semantic communication.
\end{IEEEbiography}

\begin{IEEEbiography}
[{\includegraphics[width=1in,height=1.25in,clip,keepaspectratio]{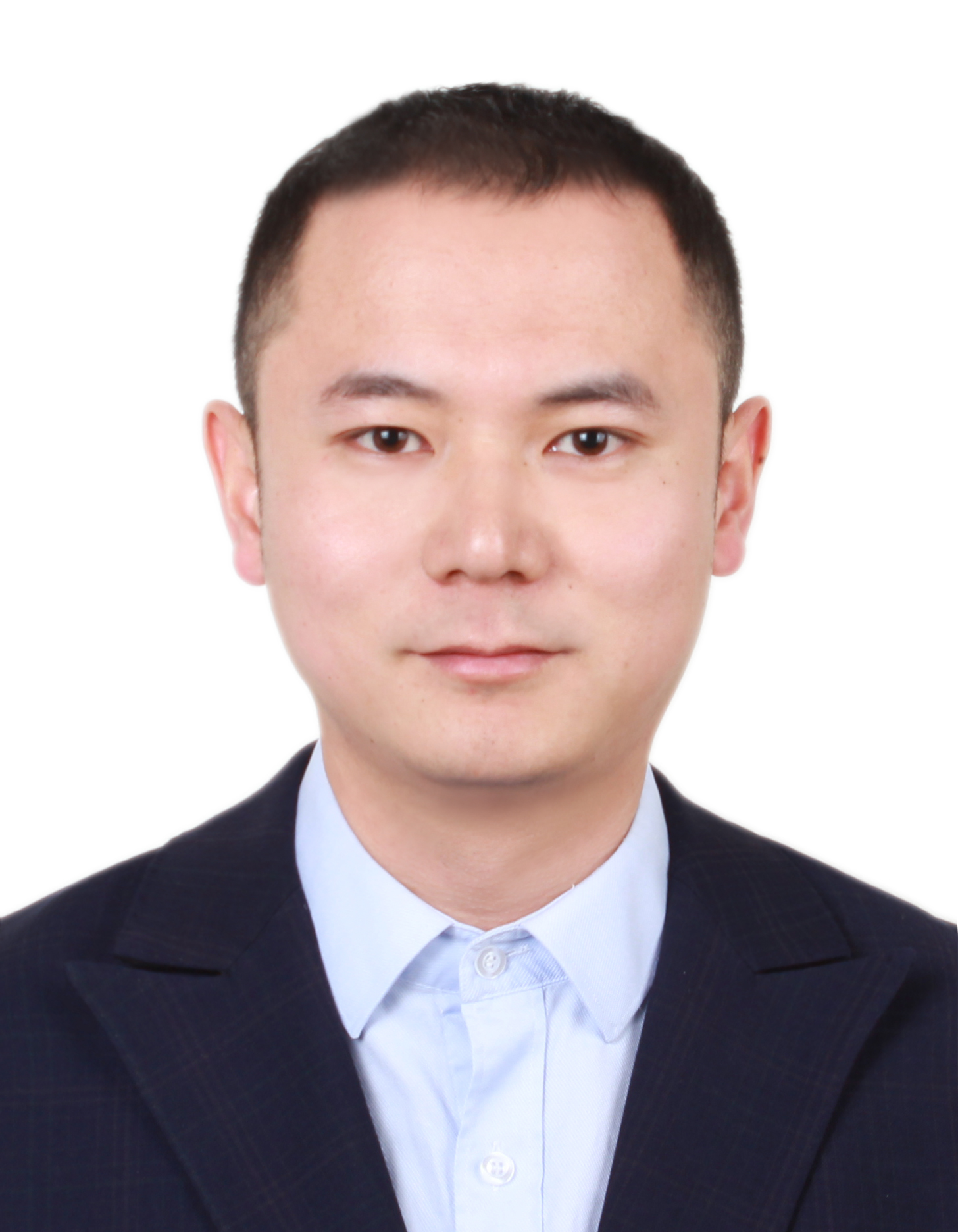}}]
{Yao Sun} (Senior Member, IEEE) is currently a Senior Lecturer with James Watt School of Engineering, the University of Glasgow, Glasgow, UK. Dr Sun has won the IEEE Communication Society of TAOS Best Paper Award in 2019 ICC, IEEE IoT Journal Best Paper Award 2022 and Best Paper Award in 22nd ICCT. His research interests include intelligent wireless networking, semantic communications, blockchain system, and resource management in next generation mobile networks. Dr. Sun is a senior member of IEEE.
\end{IEEEbiography}

\begin{IEEEbiography}
[{\includegraphics[width=1in,height=1.25in,clip,keepaspectratio]{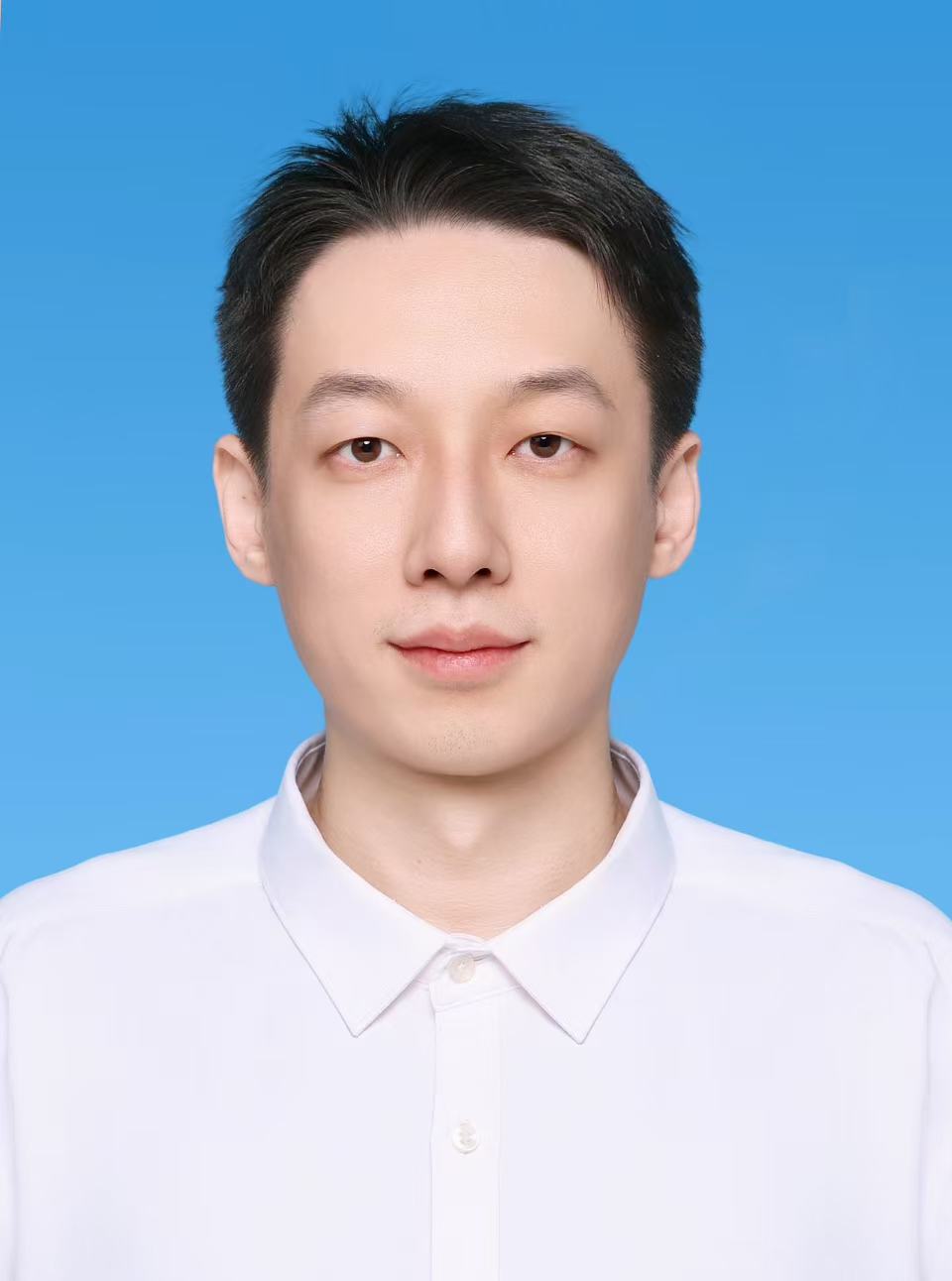}}]
{Xuesong Liu} (Graduate Student Member, IEEE) received the M.Sc. degree in Computer Science from Newcastle University in 2022. He won the Philip Merlin Best Paper Award at Newcastle University in 2022. He is currently pursuing his PhD degree at the University of Glasgow. His research interests include privacy preserving, semantic communication, and information security.
\end{IEEEbiography}

\begin{IEEEbiography}
[{\includegraphics[width=1in,height=1.25in,clip,keepaspectratio]{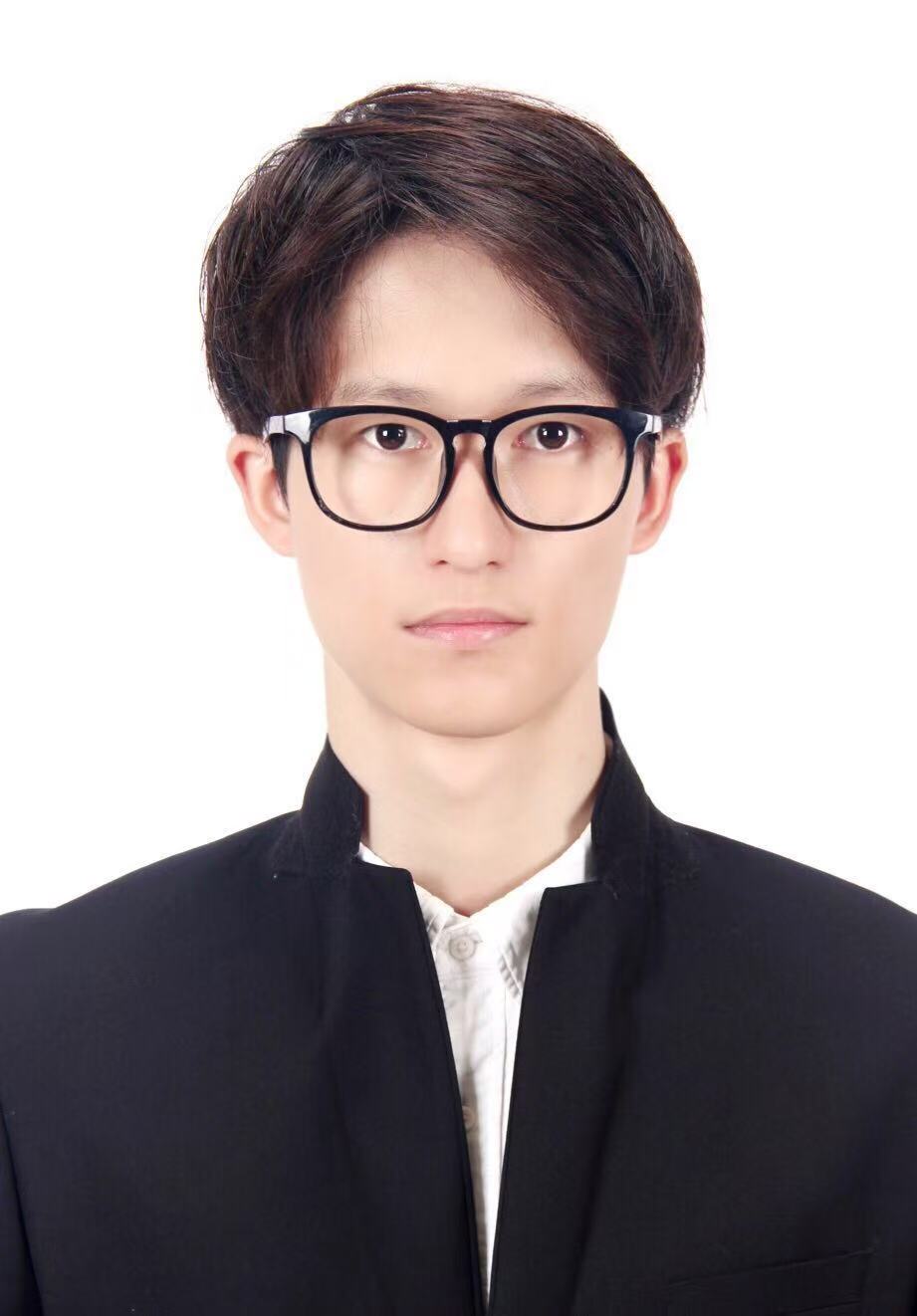}}]
{Runze Cheng} (Member, IEEE) received the Ph.D. degree in Electrical and Electronic Engineering from the James Watt School of Engineering, University of Glasgow, U.K. in 2023. He is currently a Postdoctoral Research Associate with the Communications, Sensing, and Imaging Research Group, University of Glasgow, Glasgow, U.K. His research interests include intelligent resource management, semantic communication, distributed machine learning, and space-air-ground integrated networks.
\end{IEEEbiography}

\begin{IEEEbiography}
[{\includegraphics[width=1in,height=1.25in,clip,keepaspectratio]{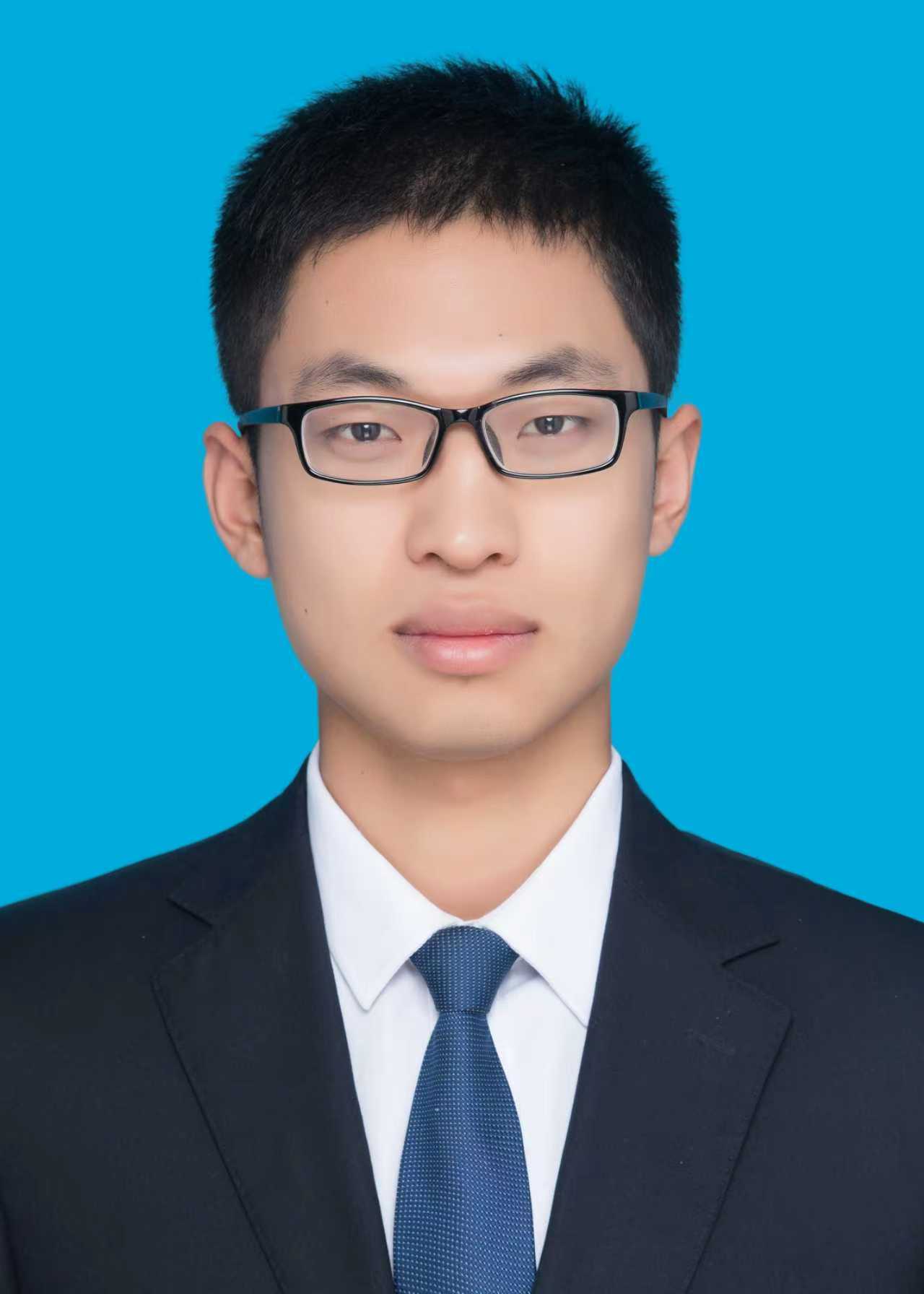}}]
{Kai Shang} received the M.S. degree in control engineering
from Beijing University of Chemical Technology, Beijing,
China, in 2017. He is a associate professor at Shandong Institute of
Petroleum and Chemical Technology, Dongying, China. He
is currently pursuing the Ph.D. degree in the School of Computer Science and
Technology, China University of Petroleum (East China),
Qingdao, China. His current research interests include image
enhancement and deep learning.
\end{IEEEbiography}

\begin{IEEEbiography}
[{\includegraphics[width=1in,height=1.25in,clip,keepaspectratio]{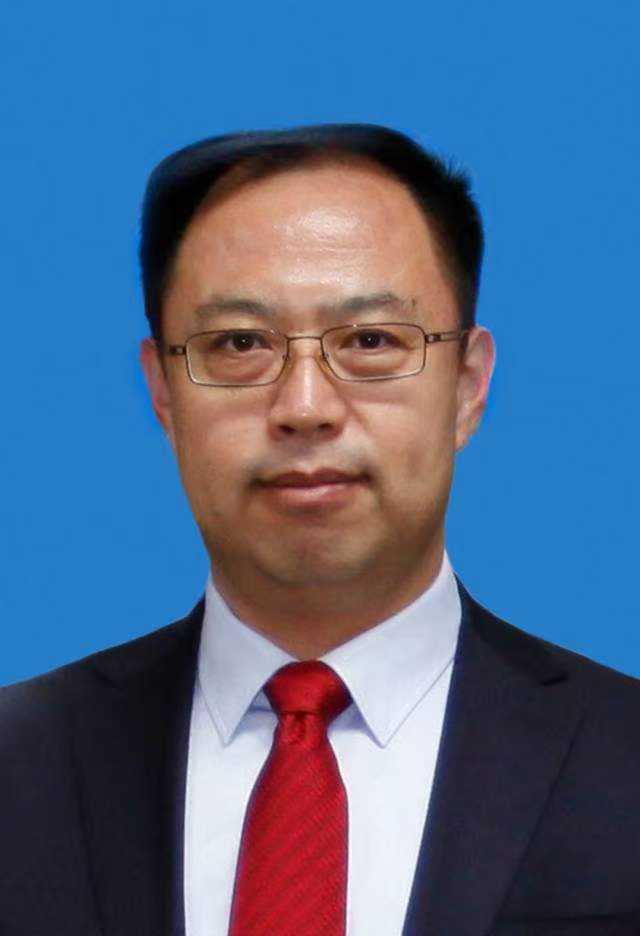}}]
{Yi Sun} was born in Liaoning, China, in 1972. He
received the M.S. degree in communication and
information system and the Ph.D. degree in electric
information technology from North China Electric
Power University, Beijing, China, in 2009 and 2014,
respectively.

He is currently a Professor of Information and
Communication Engineering with North China
Electric Power University. His mainly research
interests include smart power consumption, demand
response, and power system communication
technology
\end{IEEEbiography}
\vfill




\end{document}